\documentclass{article}

\usepackage{authblk}
\usepackage{graphicx}
\usepackage{textgreek}
\usepackage[colorlinks=true]{hyperref}
\usepackage{url}            
\usepackage{booktabs}       
\usepackage{nicefrac}       
\usepackage{amsmath,amsfonts,amssymb}
\usepackage{color}
\usepackage{microtype}      
\usepackage{xcolor}         
\usepackage{bbm}
\usepackage{algorithm}
\usepackage{algpseudocode}
\usepackage{comment}
\usepackage{xfrac}

\newcommand{\Prox}{\mathrm{Prox}}
\newcommand{\wh}[1]{{\widehat{#1}}}

\newcommand{\ov}[1]{{\overline{#1}}}
\def\ul{{\underline{r}}}
\def\ol{{\overline{r}}}
\def\Argmin{\mathop\mathrm{Argmin}}

\def\bQ{{\mathbf{Q}}}
\def\la{{\langle}}
\def\ra{{\rangle}}

\def\rank{{\hbox{\rm rank}}}
\def\card{{\mathrm{card}}}

\def\F{{\cal F}}

\def\cP{{\cal P}}

\def\bR{{\mathbf{R}}}
\def\bZ{{\mathbf{Z}}}
\def\bN{{\mathbf{N}}}

\def\L{{\cal L}}

\def\prob1{$(SO)$}

\def\B{{\cal B}}

\def\P{{\cal P}}

\def\X{{\cal X}}

\def\N{{\cal N}}

\def\bE{{\mathbf{E}}}

\def\bS{{\mathbf{S}}}
\def\Prob{\hbox{\rm Prob}}

\def\sign{\hbox{\rm sign}}

\def\qed{\hfill$\Box$}

\def\e{{\rm e}}

\newcommand{\be}{\begin{eqnarray}}
\newcommand{\ee}[1]{\label{#1}\end{eqnarray}}
\newcommand{\nn}{\nonumber \\}

\newcommand{\rf}[1]{(\ref{#1})}
\newcommand{\half}{ \mbox{\small$\frac{1}{2}$}}

\newcommand{\bse}{\begin{eqnarray*}}
\newcommand{\ese}{\end{eqnarray*}}

\definecolor{MyDarkBlue}{rgb}{0,0.08,0.45}
\definecolor{MyViolet}{rgb}{0.45,0.08,0.95}
\definecolor{MyBrown}{rgb}{0.45,0.08,0}

\newcommand{\tfo}{ \mbox{\small$\frac{3}{4}$}}

\newtheorem{theorem}{Theorem}[section]

\newtheorem{lemma}{Lemma}[section]
\newtheorem{proposition}{Proposition}[section]

\newtheorem{remark}{Remark}[section]
\newtheorem{definition}{Definition}[section]

\DeclareMathOperator*{\argmin}{arg\,min}

\usepackage{float}
\usepackage{authblk}
\usepackage{graphicx}
\usepackage{subcaption}

\def\myu{\mathfrak{r}}
\def\myU{\mathfrak{s}}

\usepackage[utf8]{inputenc}

\title{Stochastic Mirror Descent for Large-Scale Sparse Recovery}
\author{{Yannis Bekri$^{1}$ \\\vspace{-0.3cm}\href{mailto:yannis.bekri@univ-grenoble-alpes.fr}{\tt\normalsize yannis.bekri@univ-grenoble-alpes.fr}} \and 
{Sasila Ilandarideva$^{1}$\\\vspace{-0.3cm}\href{mailto:sasila.ilandarideva@univ-grenoble-alpes.fr}{\tt\normalsize sasila.ilandarideva@univ-grenoble-alpes.fr}}
\and {Anatoli Juditsky$^{1}$ \\\vspace{-0.3cm}\href{mailto:anatoli.juditsky@univ-grenoble-alpes.fr}{\tt\normalsize anatoli.juditsky@univ-grenoble-alpes.fr}}
\and {Vianney Perchet$^{2}$\\\vspace{-0.3cm}\href{mailto:vianney.perchet@normalsup.org}{\tt\normalsize vianney.perchet@normalsup.org}}\\
$^{1}$LJK, Université Grenoble Alpes, Grenoble\\
$^{2}$CREST, ENSAE Paris and CRITEO AI Lab, Paris}
\date{}

\begin{document}

\maketitle
\begin{abstract}
In this paper we discuss an application of Stochastic Approximation to statistical estimation of high-dimensional sparse parameters. The proposed solution reduces to resolving a penalized stochastic optimization problem on each stage of a multistage algorithm; each problem being solved to a prescribed accuracy by the non-Euclidean  Composite Stochastic Mirror Descent (CSMD) algorithm. Assuming that the problem objective is smooth and quadratically minorated and stochastic perturbations are sub-Gaussian, our analysis prescribes the method parameters which ensure fast convergence of the estimation error (the radius of a confidence ball of a given norm around the approximate solution). This convergence is linear during the first ``preliminary’’ phase of the routine and is sublinear during the second ``asymptotic'' phase.
We consider an application of the proposed approach to sparse Generalized Linear Regression problem. In this setting, we show that the proposed algorithm attains the optimal convergence of the estimation error under weak assumptions on the regressor distribution.  We also present a numerical study illustrating the performance of the algorithm on high-dimensional simulation data.

\end{abstract}
\section{Introduction}
Our original motivation is the well known problem of (generalized) linear high-dimensional regression with random design. Formally,  consider a dataset of $N$ points $(\phi_i, \eta_i), i \in \{1,\ldots,N\}$, where $\phi_i \in \bR^n$ are (random) features and $\eta_i \in \bR$ are observations, linked by the following equation
\be
\eta_i=\myu(\phi_i^Tx_*)+\sigma\xi_i, \quad i \in [N]:=\{1,\ldots, N\}
\ee{glr0}
where  $\xi_i\in \bR$ are i.i.d.\ observation noises. The standard objective is to recover the unknown parameter $x_*\in\bR^n$ of the Generalized Linear Regression \rf{glr0} -- which is assumed to belong to a given convex closed set $X$ and to be {\em $s$-sparse,} i.e., to have at most $s\ll n$ non-vanishing entries from the data-set.

As mentioned before,  we consider  random design, where $\phi_i$ are i.i.d.\ random variables, so that the estimation problem of $x_*$ can be recast as the following generic Stochastic Optimization problem:
\be
g_*=\min_{x\in X} g(x), \quad \text{where} \quad g(x)= \bE\big\{G\big(x,(\phi,\eta)\big)\big\},\quad G(x,(\phi,\eta))=\myU(\phi^Tx)-\phi^Tx\eta,
\ee{eq:so0}
with $\myU(\cdot)$ any primitive of $\myu(\cdot)$, i.e., $\myu(t)=\myU'(t)$. The equivalence between the original  and the stochastic optimization problems comes from the fact that $x_*$ is a critical point of $g(\cdot)$, i.e., $\nabla g(x_*)=0$ since, under mild assumptions, $\nabla g(x)=\bE\{\phi[\myu(\phi^Tx)-\myu(\phi^Tx_*)]\}$. Hence, as soon as $g$ as a unique minimizer (say, $g$  is strongly convex over $X$), solutions of both problems are identical.

As a consequence, we shall focus on the generic problem \rf{eq:so0}, that has already been widely tackled. For instance, when given an observation sample $(\phi_i,\eta_i)$, $i\in [N]$, one may build  a Sample Average Approximation (SAA) of the objective $g(x)$
\be
\wh g_N(x)=\frac{1}{N}\sum_{i=1}^N G(x,(\phi_i,\eta_i))=\frac{1}{N} \sum_{i=1}^N [\myU(\phi^T_ix)-\phi^T_ix\eta_i]
\ee{saa}
  and then solve the resulting problem of minimizing $\wh g_N(x)$ over sparse $x$'s. The celebrated $\ell_1$-norm minimization approach allows to reduce this problem to convex optimization.  We will provide a new algorithm adapted to this high-dimensional case, and   instantiating it to the original problem \ref{glr0}.

\paragraph{Existing approaches and related works.}
Sparse recovery by Lasso and Dantzig Selector has been extensively studied  \cite{candes2006stable,candes2007dantzig,bickel2009simultaneous,van2009conditions,candes2009near,candes2011probabilistic}. It computes a solution $\wh x_{N}$ to the $\ell_1$-penalized problem
$\min_{x}\wh g_N(x)+\lambda\|x\|_1$ 
where $\lambda\geq 0$ is the algorithm parameter \cite{meier2008group}.  This delivers ``good solutions'',  with high probability for sparsity level $s$ as large as $O\left(\frac{N\kappa_\Sigma}{\ln n}\right)$, as soon as  the random regressors (the $\phi_i$) are drawn independently from a normal distribution with a covariance matrix $\Sigma$ such that $\kappa_{\Sigma} I\preceq\Sigma\preceq \rho\kappa_{\Sigma}I$\footnote{We use $A\preceq B$ for two symmetric matrices $A$ and $B$ if $B-A\succeq0$, i.e. $B-A$ is positive semidefinite.}, for some $\kappa_\Sigma>0,\rho\geq 1$.
 However, computing this solution may be challenging in a very high-dimensional setting: even  popular iterative algorithms, like coordinate descent, loops over a large number of variables. To mitigate this, randomized algorithms \cite{baes2013randomized,juditsky2013randomized}, screening rules and working sets \cite{ghaoui2010safe,kowalski2011accelerating,mairal2010sparse} may be used to diminish the size of the optimization problem at hand, while iterative thresholding~\cite{aga-aos,blumensath2009iterative,jain2014iterative,Barber1,Barber2} is a ``direct'' approach to enhance sparsity of the solution.
Another approach relies on Stochastic Approximation (SA). As $\nabla G(x,(\phi_i,\eta_i))=\phi_i(\myu(\phi^T_ix)-\eta_i)$ is an unbiased estimate of $\nabla g(x)$, iterative Stochastic Gradient Descent (SGD) algorithm may  be used to build approximate solutions.
Unfortunately, unless regressors $\phi$ are sparse or possess a special structure, standard SA leads to accuracy bounds for sparse recovery  proportional to the dimension $n$ which are essentially useless in the high-dimensional setting.
This motivates non-Euclidean SA procedures, such as  Stochastic Mirror Descent (SMD) \cite{nemirovskii1979complexity}, its application  to sparse recovery enjoys almost dimension free convergence and it has been well studied in the literature.
For instance, under bounded regressors and with sub-Gaussian noise, SMD reaches ``slow rate'' of sparse recovery of the type $g(\wh x_N)-g_*=O\left({\sigma \sqrt{s\ln (n)/N}}\right)$ where $\wh x_{N}$ is the approximate solution  after
 $N$ iterations \cite{shalev2011stochastic,srebro2010smoothness}. Multistage routines may be used to improve the error estimates of SA under strong or uniform convexity assumptions \cite{juditsky2011optimization,juditsky2014deterministic,ghadimi2013optimal}. However, they do not always hold, as in sparse Generalized Linear Regression, where they are replaced by Restricted Strong Convexity conditions. Then multistage procedures \cite{agarwal2012stochastic,gaillard2017sparse} based on standard SMD algorithms \cite{juditsky2006generalization,nesterov2009primal} control the $\ell_2$-error $\|\wh{x}_N-x_*\|_2$  at the rate $O\Big(\frac{\sigma}{\kappa_\Sigma}\sqrt{\frac{s\ln n}{N}}\Big)$
with high probability. This is the best ``asymptotic'' rate attainable when solving \rf{eq:so0}.
However, those algorithms have two major limitations. They both need a number of iterations to reach a given accuracy proportional to the initial error $R=\|x_*-x_0\|_1$ and  the sparsity level $s$ must be of order $O\Big(\kappa_\Sigma\sqrt{\tfrac{N}{\ln n}}\Big)$ for the sparse linear regression. 
These limits may be seen as a consequence of dealing  with {\em non-smooth} objective $g(x)$.
Although it slightly restricts the scope of corresponding algorithms, we shall consider smooth objectives and algorithm for minimizing composite objectives (cf. \cite{juditsky2011_2,lei2018stochastic,nesterov2013gradient}) to mitigate the aforementioned drawbacks of the multistage algorithms from \cite{agarwal2012stochastic,gaillard2017sparse}.

\paragraph{Principal contributions.}
We provide a refined analysis of {\em Composite Stochastic Mirror Descent (CSMD)} algorithms for computing sparse solutions to Stochastic Optimization problem leveraging smoothness of the objective. This leads to a new ``aggressive'' choice of parameters in a multistage algorithm with significantly improved performances compared to those in \cite{agarwal2012stochastic}. We summarize below some properties of the proposed procedure for problem \rf{eq:so0}. 

Each stage of the algorithm is a specific CSMD recursion; They fall into two phases. During the first (preliminary) phase, the estimation error  decreases linearly with the exponent proportional to $\frac{\kappa_\Sigma}{s\ln n}$. When it reaches the value $O\Big(\frac{\sigma s}{\sqrt{\kappa_\Sigma}}\Big)$,  the second (asymptotic) phase begins, and its stages contain exponentially increasing number of iterations per stage, hence the estimation error decreases as $O\Big(\frac{\sigma s}{{\kappa_\Sigma}}\sqrt{\frac{\ln n}{N}}\Big)$ where $N$ is the total iteration count.

\paragraph{Organization and notation}
The remaining of the  paper is organized as follows. In Section \ref{sec:statement}, the general problem is set, and the multistage optimization routine and the study of its basic properties are presented. Then, in Section \ref{sec:sr}, we discuss the properties of the method and conditions under which it leads to ``small error'' solutions to sparse GLR estimation problems. Finally, a small simulation study illustrating numerical performance of the proposed routines in high-dimensional GLR estimation problem is presented in Section \ref{sec:num}.

In the following,  $E$ is a Euclidean space and $\|\cdot\|$ is a norm on $E$; we denote $\|\cdot\|_*$ the conjugate norm (i.e., $\|x\|_*=\sup_{\|y\|\leq 1}\la y,x\ra$).
Given a positive semidefinite matrix $\Sigma\in \bS_n$, for $x\in \bR^n$ we denote $\|x\|_\Sigma=\sqrt{x^T\Sigma x}$ and for any matrix $Q$, we denote $\|Q\|_\infty=\max_{ij}|[Q]_{ij}|$.
We use a generic notation $c$ and $C$ for absolute constants; a shortcut notation $a \lesssim b$ ($a\gtrsim b$) means that the ratio $a/b$ (ratio $b/a$) is bounded by an absolute constant; the symbols $\bigvee$,$\bigwedge$ and the notation $(.)_+$ respectively refer to "maximum between", "minimum between" and "positive part".

\section{Multistage Stochastic Mirror Descent for Sparse Stochastic Optimization}
\label{sec:statement}
This section is dedicated to the formulation of the generic stochastic optimization problem, the description and the analysis of the generic algorithm.
\subsection{Problem statement}
\label{sec:prob}
Let $X$ be a convex closed subset of an Euclidean space $E$ and $(\Omega,P)$ a probability space. We consider a mapping $G: X\times \Omega \rightarrow \bR$ such that, for all $\omega \in \Omega$,  $G(\cdot,\omega)$ is convex on $X$  and smooth, meaning that $\nabla G(\cdot,\omega)$ is Lipschitz continuous on $X$ with a.s. bounded Lipschitz constant,
\be
\forall x,x'\in X,\quad\|\nabla G(x,\omega)-\nabla G(x',\omega)\|\leq \L(\omega)\|x-x'\|,\qquad \L(\omega)\leq \nu\quad a.s..
\ee{Glip}
We define $g(x):=\bE\{G(x,\omega)\}$, where $\bE\{\cdot\}$ stands for the expectation with respect to $\omega$, drawn from $P$. We shall assume that the mapping $g(\cdot)$ is finite, convex and differentiable on $X$ and we aim at solving the following stochastic optimization problem
\begin{equation}
    \min_{x \in X} [g(x)=\bE\{G(x, \omega)\}],\label{eq:pb}
\end{equation}
assuming it admits an $s$-sparse optimal solution $x_*$ for some sparsity structure.

To solve this problem,  stochastic oracle can be queried:  when given at input a point $x \in X$, generates an $\omega\in\Omega$ from $P$ and outputs
$G(x,\omega)$ and $\nabla G(x,\omega):=\nabla_xG(x,\omega)$ (with a slight abuse of notations). We assume that the oracle is {\em unbiased}, i.e.,
\[
\bE\{\nabla G(x,\omega)\}=\nabla g(x),\qquad \forall x\in X.
\]

To streamline presentation, we assume, as it is often the case in  applications of stochastic optimization problem \rf{eq:pb}, that $x_*$ is unconditional, i.e., $\nabla g(x_*)=0$.
or stated otherwise $\bE\{\nabla G(x_*,\omega)\}=0$;
we also suppose the sub-Gaussianity of $\nabla G(x_*,\omega)$,  namely that,
for some $\sigma_*<\infty$
\be
\bE\Big\{\exp\Big({\|\nabla G(x_*,\omega)\|^2_*/\sigma_*^2}\Big)\Big\}\leq \exp(1).
\ee{eq:subgaus}
\subsection{Composite Stochastic Mirror Descent algorithm}
\label{sec:CSMD}
As mentioned in the introduction, (stochastic) optimization over the set of sparse solutions can be done through "composite" techniques. We take a similar approach here, by transforming the generic problem  \ref{eq:pb} into  the following {\em composite Stochastic Optimization problem}, adapted to some norm   $\|\cdot\|$, and parameterized by $\kappa\geq 0$,
\begin{equation}
    \min_{x \in X} \big[F_\kappa(x) := \half g(x) + \kappa\|x\|=\half \bE\{G(x, \omega)\}+\kappa\|x\|\big].\label{eq:comp}
\end{equation}
The purpose of this section is to derive a new (proximal) algorithm. We first provide necessary backgrounds and notations.

\paragraph{Proximal setup, Bregman divergences and Proximal mapping.}
Let $B$ be the unit ball of the norm $\|\cdot\|$ and  $\theta:\,B\to \bR$ be a {\em distance-generating function (d.-g.f.)} of $B$, i.e., a continuously differentiable  convex function which is strongly convex with respect to the norm $\|\cdot\|$,
\[
\la\nabla \theta(x)-\nabla \theta(x'),x-x'\rangle\ge \|x-x'\|^2,\quad \forall x,x'\in X.
\]
We  assume w.l.o.g. that $\theta(x)\geq \theta(0)=0$ and denote $\Theta=\max_{\|z\|\leq 1}\theta(z)$.

We now introduce a local and renormalized version of the d.-g.f.\ $\theta$.

\begin{definition}For any $x_0 \in X$, let $X_R(x_0):= \{z\in X: \|z-x_0\|\leq R\}$ be the ball of radius $R$ around $x_0$. It is equipped   with the d.-g.f. $\vartheta^R_{x_0}(z) := R^2\theta\left((z-x_0)/R\right)$.
\end{definition}
Note that $\vartheta^R_{x_0}(z)$ is strongly convex on ${X}_R(x_0)$ with modulus 1,  $\vartheta^R_{x_0}(x_0)=0$, and $\vartheta^R_{x_0}(z)\leq \Theta R^2$.

\begin{definition} Given $x_0 \in X$ and $R>0$,  the Bregman divergence $V$ associated to $\vartheta$  is defined by	\[
	V_{x_0}(x,z)=\vartheta^R_{x_0}(z)-\vartheta^R_{x_0}(x)-\la\nabla \vartheta^R_{x_0}(x),z-x\ra,\quad x,z\in X.
	\]
\end{definition}

We can now define {\em composite proximal mapping}  on $X_R(x_0)$ \cite{nesterov2013gradient,nesterov2013first} with respect to some convex and continuous mapping $h:\,X\to \bR$. \begin{definition}
The  composite proximal mapping with respect to $h$ and $x$ is defined by
\be
\Prox_{h,x_0}(\zeta,x)&:=&\argmin_{z\in {X_R(x_0)}}\big\{\langle  \zeta,z\rangle +h(z)+V_{x_0}(x,z)\big\}\nn
 &=&\argmin_{z\in {X_R(x_0)}}\big\{\langle  \zeta-\nabla\vartheta^R_{x_0}(x),z\rangle +h(z)+\vartheta^R_{x_0}(z)\big\}
	\ee{cprox}
\end{definition}
If \rf{cprox} can be efficiently solved to high accuracy and $\Theta$ is ``not too large'' (we refer to \cite{juditsky2011optimization,nemirovski2009robust,nesterov2013first}); those setups will be called ``prox-friendly''. 
We now introduce the main building block of our algorithm, the Composite Stochastic Mirror Descent.

\paragraph{Composite Stochastic Mirror Descent algorithm.}
Given a sequence of positive {\em step sizes} $\gamma_i>0$, the {\em Composite Stochastic Mirror Descent} (CSMD) is defined by the following recursion
\be
x_{i}&=\Prox_{\gamma_ih,x_0}(\gamma_{i-1}  \nabla G(x_{i-1},\omega_i),x_{i-1}),\quad x_0\in X.
\ee{eq:fa}
After  $m$ steps of CSMD, the final output is  $\wh x_m$ (approximate solution) defined by
\be
\wh x_m =\frac{\sum_{i=0}^{m-1} \gamma_ix_i}{\sum_{i=0}^{m-1}\gamma_i}
\ee{eq:asol}
For any integer $L\in \bN$, we can also define the $L$-minibatch CSMD. Let $\omega_i^{(L)}=[\omega_i^1,...,\omega_i^L]$ be i.i.d.\ realizations of $\omega_i$. The associated (average) stochastic gradient is then simply defined as
\[
H\left(x_{i-1},\omega^{(L)}_i\right)={1\over L}\sum_{\ell=1}^L \nabla G(x_{i-1},\omega^{\ell}_i),
\]
which yields the following recursion for the $L$-minibatch CSMD recursion:
 \be
x_{i}^{(L)}&=\Prox_{\gamma_ih,x_0}\left(\gamma_{i-1}  H\left(x_{i-1},\omega^{(L)}_i\right),x_{i-1}^{(L)}\right),\quad x_0\in X,
\ee{eq:fa_mini}
with its  approximate solution $\wh x_m^{(L)} =\sum_{i=0}^{m-1} \gamma_ix_i^{(L)}/\sum_{i=0}^{m-1}\gamma_i$ after $m$ iterations.

From now on, we set $h(x)=\kappa \|x\|$.
\begin{proposition}\label{pr:mupropn}
If step-sizes are constant, i.e., $\gamma_i\equiv \gamma\le (4\nu)^{-1}$, $i=0,1,...$, and the initial point $x_0\in X$ such that $x_*\in X_R(x_0)$ then for any $t\gtrsim \sqrt{1+\ln m}$, 
with probability at least $1-4e^{-t}$
\be
F_\kappa(\wh x_m)-F_\kappa({x_*})\lesssim m^{-1}\big[\gamma^{-1}R^2(\Theta+t)+\kappa R+\gamma\sigma^2_*(m+t)\big],
\ee{eq:almo}
and the approximate solution $\wh x_m^{(L)}$ of the $L$-minibatch CSMD satisfies
\be
F_\kappa(\wh x_m^{(L)})-F_\kappa({x_*})\lesssim m^{-1}\big[\gamma^{-1} R^2(\Theta+t)+\kappa R+\gamma\sigma^2_*\Theta L^{-1}(m+t)\big].
\ee{eq:almo2}
	\end{proposition}

For the sake of clarity and conciseness, we denote CSMD($x_0, \gamma, \kappa, R, m, L$) the approximate solution $\widehat{x}^{(L)}_m$ computed after $m$ iterations of $L$-minibatch CSMD algorithm with initial point $x_0$, step-size $\gamma$, and radius $R$ using recursion \eqref{eq:fa_mini}.

\subsection{Main contribution: a multistage adaptive algorithm}
\label{sec:multi_alg}
Our approach to find sparse solution to the original stochastic optimization problem  \rf{eq:comp} consists in solving a  sequence of auxiliary composite problems \rf{eq:comp}, with their sequence of parameters ($\kappa$, $x_0$, $R$) defined recursively.  For the latter, we need to infer the quality of approximate solution to \rf{eq:pb}. To this end, we introduce the following {\em Reduced Strong Convexity} (RSC) assumption, satisfied in the motivating example (it is discussed in the appendix for the sake of fluency):

\paragraph{Assumption [RSC]}
There exist some $\delta>0$  and $\rho<\infty$ such that for any feasible solution $\widehat{x} \in X$ to the composite problem \rf{eq:comp} satisfying, with probability at least $1-\varepsilon$,
\[
    F_{\kappa}(\widehat{x}) - F_\kappa(x_*) \leq \upsilon,
\]
it holds, with probability at least $1-\varepsilon$, that 
\be
\|\wh x - x_*\|\leq \delta\left[\rho s\kappa + {\upsilon \kappa^{-1}}\right].
\ee{eq:lemquad}

Given the different problem parameters $s, \nu, \delta, \rho, \kappa, R$ and some initial point $ x_0 \in X$ such that $x_* \in X_R(x_0)$ Algorithm \ref{alg2} works in stages. Each stage represents a  run of CSMD algorithm with properly set penalty parameter $\kappa$. More precisely, at stage $k+1$, given the approximate  solution $\widehat{x}^k_{m}$ of stage $k$, a new instance of CSMD is initialized on $X_{R_{k+1}}(x^{k+1}_0)$  with $x^{k+1}_0=\widehat{x}^k_{m}$ and $R_{k+1}=R_k/2$.

Furthermore, those stages are divided into two phases which we refer to as {\em preliminary} and {\em asymptotic}:
\begin{description}
\item[Preliminary phase:] During this phase, the  step-sizes $\gamma$ and the number of CSMD iterations per stage are fixed; the error of approximate solutions converges linearly with the total number of calls to stochastic oracle. This phase terminates when the error of approximate solution becomes independent of the initial error of the algorithm; then the asymptotic phase begins.
\item[Asymptotic phase:] In this phase, the step-size decreases and the length of the stage increases linearly; the solution  converges sublinearly, with the ``standard'' rate $O\big(N^{-1/2}\big)$ where $N$ is the total number of oracle calls. When expensive proximal computation \rf{cprox} results in high numerical cost of the iterative algorithm, minibatches are used to keep  the number of iterations per stage fixed.
\end{description}

\begin{algorithm}[h]
\caption{CSMD-SR}\label{alg2}
\textbf{Initialization :} Initial point $x_0\in X$, step-size $\gamma = (4\nu)^{-1}$, initial radius $R_0$, confidence level $t$, total budget $N$.
\\ Set $m_0\asymp{s\rho\nu\delta^2 (\Theta + t)}$, $\overline{K}_1 \asymp \ln\left({R_0^2 \nu\over \delta^2\rho\sigma^2_*s}\right)\wedge {N\over m_0}$, $L =1$
\begin{algorithmic}
\State {\textbf{if} $R_0\gtrsim \sigma_*\delta\sqrt{\rho s\over \nu}$ \textbf{continue} with preliminary stage,
\State\textbf{else} proceed directly to asymptotic phase
\State \textbf{end}}
\For{\textbf{stage} $k=1,\dots, \overline{K}_1 $}\Comment{\underline{Preliminary Phase}}
\State Set $\kappa_k \asymp R_{k}(\delta\rho s)^{-1}$
\State Compute approximate solution $\widehat{x}^k_{m_0}=$CSMD($x_0, \gamma, \kappa_k, R_k, m_0, L$) at stage $k$
\State Reset the prox-center $x_0 = \widehat{x}^k_{m_0}$
\State Set $R_k = R_{k-1}/2$
\EndFor
\State Set $\widehat{x}_N = \widehat{x}_{m_0}^{\overline{K}_1}$, $B = N - m_0 \overline{K}_1 $, $ m_1 \asymp m_0$
\State\textbf{if} {$m_1 >B$} \textbf{output :} $\widehat{x}_N$ and \textbf{return}; \textbf{endif}\Comment{\underline{Asymptotic Phase}}
\State Set $r_0 = R_{\overline{K}_1}$
\State Set $k = 1$
\While{$m_k \leq B$}
\State Set $\kappa_k \asymp 2^{-k}\sigma_*(\rho\nu s)^{-1/2}$, $\gamma_k\asymp 4^{-k}\nu^{-1}$
\State Compute approximate solution $\widehat{x}^k_{m_k}=$CSMD($x_0, \gamma_k , \kappa_k, r_k, m_k, L$) at stage $k$
\State Reset the prox-center $x_0  = \widehat{x}^k_{m_k}$
\State Set $B = B - m_k$, $k=k+1$, $r_k = r_{k-1}/2$, $ m_k \asymp 4^k m_0$
\EndWhile
\end{algorithmic}
\textbf{output :} $\widehat{x}_N = \widehat{x}^{k}$
\end{algorithm}
%
In the algorithm description, $\overline{K}_1$ and $\overline{K}_2 \asymp 1+\log(\frac{N}{m_0})$ stand for the respective maximal number of stages of the two phases of the method, here, $m_0\asymp{s\rho\nu\delta^2 (\Theta + t)}$ is the length of stages of the first (preliminary) phase. The pseudo-code for the variant of the asymptotic phase with minibatches is given in Algorithm \ref{alg2-minibatch}.
\begin{algorithm}[h]
\caption{Asymptotic phase of CSMD-SR with minibatch}\label{alg3}
\textbf{Input :} The approximate solution $\widehat{x}^{\overline{K}_1}_{m_0}$ at the end of the preliminary stage, step-size parameter $\gamma$, radius at the end of the preliminary phase $R_{\overline{K}_1}$, initial batch size $\ell_1\asymp\Theta$
\begin{algorithmic}[1]
\State Set $r_0 = R_{\overline{K}_1}$, $x_0=\widehat{x}^{\overline{K}_1}_{m_0}$, $B = N - m_0 \overline{K}_1 $\Comment{\underline{Asymptotic Phase}}
\State $k = 1$
\While{$m_0 \ell_k \leq B$}
\State  $\kappa_k \asymp 2^{-k}\sigma_*(\rho\nu s)^{-1/2}$
\State  Compute approximate solution $\widehat{x}^k_{m_0}=$CSMD($x_0, \gamma_k, \kappa_k,  r_k, m_0, L=\ell_k$) at stage $k$
\State Reset the prox-center $x_0  = \widehat{x}^k_{m_0}$
\State Set $B = B - m_0 \ell_k$, $k=k+1$, $r_k = r_{k-1}/2$, $ \ell_k \asymp 4^k\ell_1$
\EndWhile
\end{algorithmic}
\textbf{output:} $\widehat{x}^{(b)}_N= \widehat{x}^{k}_{m_{2}}$ \label{alg2-minibatch}

\end{algorithm}

The following theorem states the main result of this paper, an upper bound on the precision of the estimator computed by our multistage method.
\begin{theorem}\label{thm:reli}
Assume that the total sample budget satisfies $N \geq m_0$, so that at least one stage of the preliminary phase of Algorithm \ref{alg2} is completed, then for $ t\gtrsim \sqrt{\ln N}$ the  approximate solution $\widehat{x}_N$ of Algorithm \ref{alg2} satisfies, with probability at least $1 - C(\overline{K}_1 + \overline{K}_2)e^{-t}$,
\[
    \|\widehat{x}_N - x_*\| \lesssim R\exp\left\{-\frac{c}{{\delta^2} \rho\nu}{N\over   s(\Theta + t)}\right\} + {{ \delta^2}\rho\sigma_* s}\sqrt{\frac{ \Theta + t}{N}}.
\]
The corresponding solution $\widehat{x}^{(b)}_N$ of the minibatch Algorithm \ref{alg3} satisfies with  probability $\geq 1 - C(\overline{K}_1 + \widetilde{K}_2)e^{-t}$
\[
    \|\widehat{x}^{(b)}_N - x_*\| \lesssim R\exp\left\{-\frac{c}{{\delta^2}\rho \nu } { N\over s\left(\Theta + t\right)}\right\} + {{ \delta^2}\rho\sigma_*s}{}\sqrt{\frac{\Theta \left(\Theta + t\right)}{N}}.
\]
where $\widetilde{K}_2\asymp 1+\ln\big(\frac{N}{\Theta m_0}\big)$ is the bound for the number of stages of the asymptotic phase of the minibatch algorithm.
\end{theorem}

\begin{remark}
Along with the oracle computation, proximal computation to be implemented at each iteration of the algorithm is an important part of the computational cost of the method. It becomes even more important during the asymptotic phase when number of iterations per stage increases exponentially fast with the stage count, and may result in poor real-time convergence. The interest of  minibatch implementation of the second phase of the algorithm is in reducing drastically the number of iterations per asymptotic stage.  The price to  paid is an extra factor $\sqrt{\Theta}$ that could also theoretically hinder convergence. However, in the problems of interest (sparse and group-sparse recovery, low rank matrix recovery) $\Theta$ is logarithmic in problem dimension. Furthermore, in our numerical experiments we did not observe any accuracy degradation when using the minibatch variant of the method.
\end{remark}

\section{Sparse generalized linear regression by stochastic approximation}\label{sec:sr}
\subsection{Problem setting}\label{sec:rps}
We now consider again the original  problem of recovery of a $s$-sparse signal $x_*\in  X\subset\bR^n$ from random observations defined by 
\be
\eta_i=\myu(\phi_i^Tx_*)+\sigma\xi_i,\;\;\;i=1,2,...,N,
\ee{sparselin}
where $\myu:\bR\to \bR$ is some non-decreasing and continuous ``activation function'', and $\phi_i\in \bR^{n}$ and $\xi_i\in \bR$ are mutually independent.
We assume that $\xi_i$ are sub-Gaussian, i.e., $\bE\big\{e^{\xi_i^2}\big\}\leq \exp(1)$,
 while regressors $\phi_{i}$ are bounded, i.e.,
$
\|\phi_i\|_\infty\leq \ov{\nu}$. We also denote $\Sigma=\bE\{\phi_{i}\phi_{i}^T\}$, with 
$\Sigma\succeq\kappa_\Sigma I$ with some $\kappa_\Sigma>0$, and $\|\Sigma_j\|_\infty\leq \upsilon<\infty$.

We will apply the machinery developed in Section \ref{sec:statement}, with respect to
\[g(x)=\bE\big\{\myU(\phi^Tx)-x^T\phi\eta \big\}
\]
where $\myu(t)=\nabla \myU(t)$ for some convex and continuously differentiable $\myU$, applied with  the norm $\|\cdot\|=\|\cdot\|_1$ (hence $ \|\cdot\|_*=\|\cdot\|_\infty$), from some initial point $x_0\in X$ such that $\|x_*-x_0\|_1\leq R$. It remains to prove that the different assumptions of Section \ref{sec:statement} are satisfied.

\begin{proposition}\label{PR:mild}
Assume that  $\myu$ is $\ol$-Lipschitz continuous and $\ul$-strongly monotone (i.e., $|\myu(t)-\myu(t')|\geq \ul|t-t'|$ which implies that $\myU$ is $\ul$-strongly convex) then
\begin{enumerate}
\item {[Smoothness]} $G(\cdot,\omega)$ is $\mathcal{L}(\omega)$-smooth with $\mathcal{L}(\omega)\leq \ol \ov\nu^2.$
    \item {[Quadratic minoration]} $g$ satisfies
    \be
    g(x)-g(x_*)\geq \half \ul\|x-x_*\|^2_\Sigma.
    \ee{eq:qmin}
   \item {[Reduced Strong Convexity]} Assumption [RSC] holds with { $\delta=1$} and $\rho=(\kappa_\Sigma\ul)^{-1}$.
    \item {[Sub-Gaussianity]} $\nabla G(x_*,\omega_i)$ is $\sigma^2\ov\nu^2$-sub Gaussian.
 \end{enumerate}
\end{proposition}
The proof is postponed to the appendix. The last point is a consequence of a generalization of the Restricted Eigenvalue property \cite{bickel2009simultaneous}, that we detail below (as it gives insight on why Proposition  \ref{PR:mild} holds).

This condition, that we state and call $\bQ(\lambda, \psi)$ in the following Lemma \ref{lem:3.1sparse}, and  is reminiscent of \cite{juditsky2011accuracy} with the corresponding assumptions of \cite{raskutti2010restricted,dalalyan2019outlier}.

\begin{lemma} \label{lem:3.1sparse}
Let $\lambda>0$ and $0<\psi\leq 1$, and suppose that for all subsets $I\subset \{1,...,n\}$ of cardinality smaller than $s$ the following property is verified:
\begin{align}
\forall z\in \bR^n\quad\|z_I\|_1\leq  \sqrt{s\over \lambda} \|z\|_{\Sigma}+\half(1-\psi)\|z\|_1\tag*{$\bQ(\lambda, \psi)$}
\label{z_I}
\end{align}
where  $z_I$ is obtained by zeroing all its components with indices $i \notin I$.

If $g(\cdot)$ satisfies the {\em quadratic minoration condition}, i.e., for some $\mu>0$,
\be
g(x)-g(x_*)\geq \half \mu\|x-x_*\|_{\Sigma}^2,
\ee{qminc}
and that $\wh x$ is an admissible solution to \eqref{eq:comp} satisfying, with probability at least $1-\varepsilon$,
\[
F_\kappa(\wh x)\leq F_\kappa(x_*)+\upsilon.
\]Then, with probability at least $1-\varepsilon$,
\be
\|\wh x-x_*\|_1\leq {s\kappa\over \lambda\mu\psi}+ {\upsilon\over \kappa\psi}.
\ee{quadsp}
\end{lemma}

\begin{remark}
Condition \ref{z_I} generalizes the classical Restricted Eigenvalue (RE) property \cite{bickel2009simultaneous} and Compatibility Condition  \cite{van2009conditions}, and is the most relaxed condition under which classical bounds for the error of $\ell_1$-recovery routines were established. Validity of \ref{z_I} with some $\lambda>0$ is necessary for $\Sigma$ to possess the celebrated {\em null-space property}  \cite{cohen2009compressed}
\[
\exists \psi>0:\; \max_{I,\,|I|\leq s}\|z_I\|_1\leq \half (1-\psi) \|z\|_1\;\;\forall z\in\mathrm{Ker}(\Sigma)
\]
which is necessary and sufficient for the $s$-goodness of $\Sigma$ (i.e., $\widehat{x}\in\Argmin_{u}\left\{\|u\|:\; \Sigma u=\Sigma x_*\right\}$ reproduces exactly  every $s$-sparse signal $x_*$ in the noiseless case).

When $\Sigma$ possesses the nullspace property, \ref{z_I} may hold for $\Sigma$ with nontrivial kernel; this is typically the case for random matrices \cite{raskutti2010restricted,rauhut2010compressive} such as rank deficient Wishart matrices, etc. When $\Sigma$ is a regular matrix, condition \ref{z_I} may also holds with constant $\lambda$ which is much higher that the minimal eigenvalue of $\Sigma$ when the eigenspace corresponding to small eigenvalues of $\Sigma$ does not contain vectors $z$ with $\|z_I\|_1> \half(1-\psi)\|z\|_1$.
\end{remark}


\paragraph{Remarks.}
In the case of linear regression where $\myu(t)=t$, it holds
\bse
g(x)&=&\bE\big\{\half(\phi^Tx)^2-x^T\phi\eta\big\}=\half \bE\big\{(\phi^T(x_*-x))^2-(\phi^Tx_*)^2\big\}\\&=&
\half (x-x_*)^T\Sigma(x-x_*)-\half x_*^T\Sigma x_*=
\half \|x-x_*\|_\Sigma^2-\half \|x_*\|_\Sigma^2
\ese
and
$\nabla G(x,\omega)=\phi\phi^T(x-x_*)-\sigma\xi\phi$.
In this case
$
\L(\omega)\leq \|\phi\phi^T\|_\infty\leq \ov\nu^2.
$

\begin{figure}[h]
\centering
\begin{center}
\begin{tabular}{cc}
\includegraphics[width=0.3\textwidth]{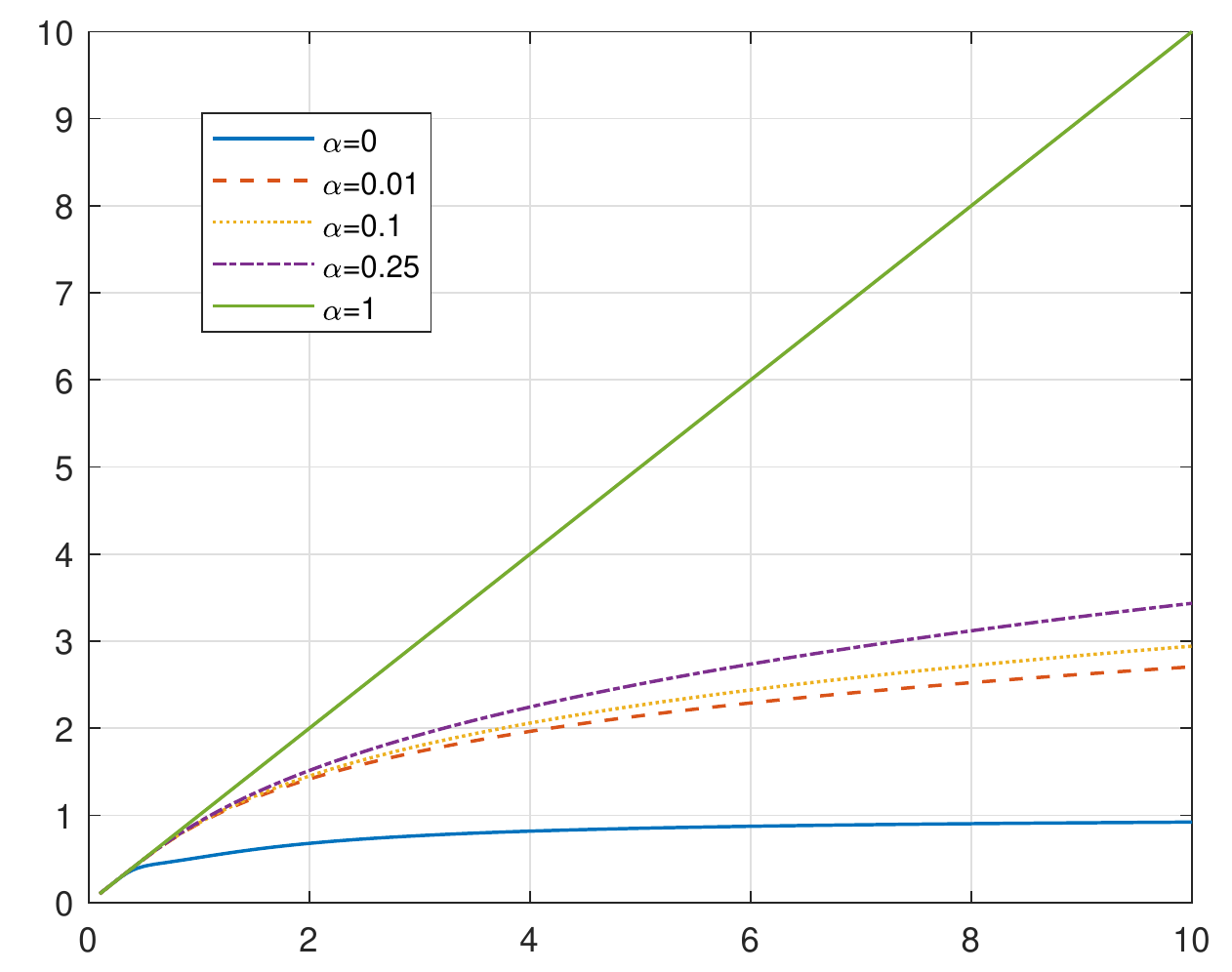}\quad&\quad\includegraphics[width=0.3\textwidth]{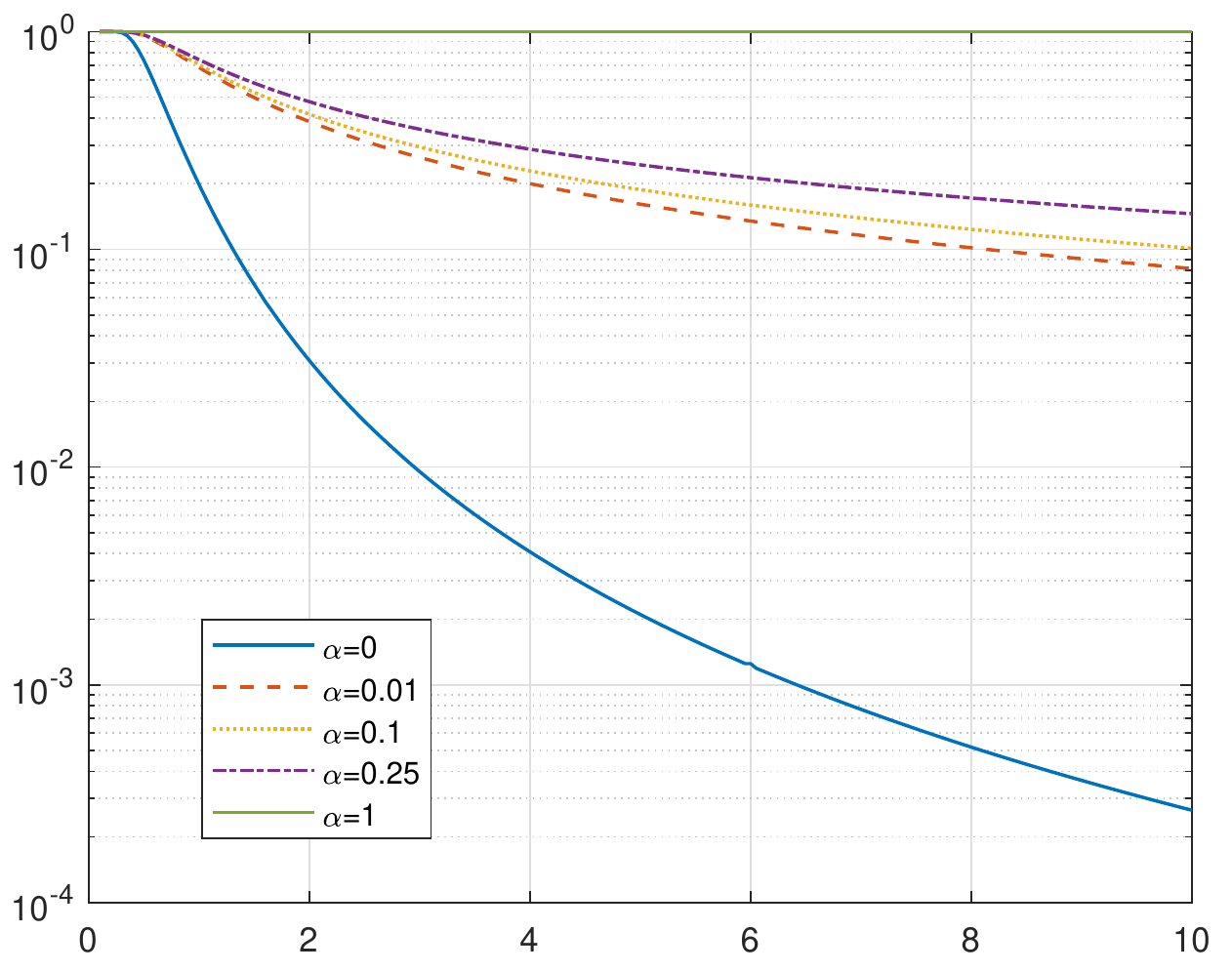}\\
\end{tabular}
\end{center}
\caption{\small\label{fig:funplot} Given the activation function $\myu$ in \rf{ral} and $\alpha=(0,0.01,0.1,0.25,1)$; left plot:  mappings $h$; right plot: moduli of strong monotonicity of  mappings $H$ on $\{x:\|x\|_\Sigma\leq r\}$ as function of $r$.}
 \end{figure}

Note that quadratic minoration bound \rf{eq:qmin} for $g(x)-g(x_*)$ is often overly pessimistic. Indeed,  consider for instance, Gaussian regressor $\phi\sim \N(0,\Sigma)$ (such regressors are not a.s. bounded, we consider this example only for illustration purposes) and activation $\myu$, define for some $0\leq \alpha\leq 1$ (with the convention, $0/0=0$)
\be
\myu(t)=\left\{\begin{array}{ll}t,&|t|\leq 1,\\
\sign(t)[\alpha^{-1}(|t|^\alpha-1)+1],&|t|>1. \end{array}
\right.
\ee{ral}
When passing from $\phi$ to $\varphi=\Sigma^{-1/2}\phi$ and from $x$ to $z=\Sigma^{1/2}x$ and using the fact
that
\[
\varphi={zz^T\over \|z\|_2^2}\varphi+\underbrace{\left(I-{zz^T\over \|z\|_2^2}\right)\varphi}_{=:\chi}
\]
with independent ${zz^T\over \|z\|_2^2}\varphi$ and $\chi$, we obtain
\[
H(x)=\bE\{\phi[\myu(\phi^Tx)]\}=\bE\left\{{zz^T\over \|z\|_2^2}\varphi\,\myu(\varphi^Tz)\right\}={z\over \|z\|_2}\bE\left\{\varsigma\myu(\varsigma\|z\|_2)\right\}=
{\Sigma^{1/2}x\over \|x\|_\Sigma}\bE\left\{\varsigma\myu(\varsigma\|x\|_\Sigma)\right\}
\]
where $\varsigma\sim \N(0,1)$. Thus, $H(x)$ is proportional to ${\Sigma^{1/2}x\over \|x\|_\Sigma}$ with coefficient
 \[h\big(\|x\|_\Sigma\big)=\bE\left\{\varsigma\myu(\varsigma\|x\|_\Sigma)\right\}.
  \]
  Figure \ref{fig:funplot} represents the mapping $h$ for different values of $\alpha$ (on the left), along with the corresponding mapping $H$ on a $\|\cdot\|_\Sigma$-ball centered at the origin  of radius $r$ (on the right).

\subsection{Stochastic Mirror Descent algorithm}
In this section, we  describe the statistical properties of approximate solutions of Algorithm \ref{alg2} when applied to the sparse recovery problem.  We shall use the following distance-generating function of the $\ell_1$-ball of $\bR^n$ (cf. \cite[Section 5.7.1]{juditsky2011optimization})
\be
    \theta(x) = {c \over p} \|x\|_p^p, \quad p = \left\{\begin{array}{ll}2,& n=2\\1 + \frac{1}{\ln(n)},&n\geq 3,\end{array}\right.\quad c=\left\{\begin{array}{ll}2,& n=2,\\e\ln n,&n\geq 3.\end{array}\right. \label{proxfct}
\ee{thel1}
It immediately follows that $\theta$ is strongly convex with modulus 1 w.r.t.\ the norm $\|\cdot\|_1$ on its unit ball, and that $\Theta\leq   e \ln n$. In particular, Theorem \ref{thm:reli} entails the following statement.
\begin{proposition}\label{cor:mycorL1}  For $ t\gtrsim\sqrt{\ln N}$, assuming the samples budget is large enough, i.e., $N \geq m_0$ (so that at least one stage of the preliminary phase of Algorithm \ref{alg2} is completed),  the approximate solution $\widehat{x}_N$ output  satisfies  with probability at least $1 - Ce^{-t}\ln N$,
\be
    {\|\widehat{x}_N - x_*\|_1} \lesssim R\exp\left\{-c\frac{\ul\kappa_\Sigma}{\ol\ov\nu^2}{N\over  s({\ln n} + t)}\right\} + \frac{\sigma\ov\nu s}{\ul\kappa_\Sigma}\sqrt{\frac{ {\ln n} + t}{N}}
\ee{eq:prop32nob}
The corresponding solution $\widehat{x}^{(b)}_N$ of the minibatch variant of the algorithm satisfies with  probability $\geq 1 - Ce^{-t}\ln N$, 
\[
   {\|\widehat{x}_N^{(b)} - x_*\|_1} \lesssim R\exp\left\{-c\frac{\ul\kappa_\Sigma}{ \ol\ov\nu^2 } { N\over s\left({\ln n} + t\right)}\right\} + \frac{\sigma\ov\nu
    s}{\ul\kappa_\Sigma}\sqrt{\frac{{\ln n} \left({\ln n} + t\right)}{N}}
\]
\end{proposition}

\begin{remark}
Bounds for the $\ell_1$-norm of the error $\wh x_N-x_*$ (or $\widehat{x}_N^{(b)}-x_*$) established in Proposition \ref{cor:mycorL1} allows us to quantify prediction error $g(\wh x_N)-g(x_*)$ (and $g(\wh x^{(b})_N)-g(x_*)$, and also lead to bounds for $\|\wh x_N-x_*\|_\Sigma$ and $\|\wh x_N-x_*\|_2$ (respectively, for $\|\wh x_N^{(b)}-x_*\|_\Sigma$ and $\|\wh x_N^{(b)}-x_*\|_2$).
For instance, Proposition \ref{pr:mupropn} in the present setting implies the bound on the prediction error after $N$ steps of the algorithm that reads
\begin{align*}
g(\wh x_N)-g(x_*)\lesssim {R^2\kappa_\Sigma \ul\over s}\exp\left\{-\frac{c\kappa_\Sigma \ul}{\delta^2 \ol\ov\nu^2}{N\over   s(\Theta + t)}\right\} +{\sigma^2\ov\nu^2s (\Theta+t)\over \kappa_\Sigma \ul N}
\end{align*}
with probability $\geq 1-C\ln N e^{-t}$. We conclude by \rf{eq:qmin} that
\begin{align*}
&\|\wh x_N-x_*\|_2^2\leq \kappa_\Sigma^{-1}\|\wh x_N-x_*\|^2_\Sigma \leq 2\kappa_\Sigma^{-1}\ul^{-1} [g(\wh x_N)-g(x_*)]\\
&\quad\lesssim {R^2\over s}\exp\left\{-\frac{c\kappa_\Sigma \ul}{\delta^2 \ol\ov\nu^2}{N\over   s(\Theta + t)}\right\}+{\sigma^2\ov\nu^2s (\Theta+t)\over \kappa^2_\Sigma \ul^2 N}.
\end{align*}

In other words, the error $\|\wh x_N-x_*\|_2$  converges geometrically to the ``asymptotic rate'' ${\sigma\ov\nu\over \kappa_\Sigma \ul}\sqrt{s (\Theta+t)\over  N}$ which is the ``standard'' rate established in the setting (cf. \cite{aga-aos,bickel2009simultaneous,meier2008group}, etc).
\end{remark}
\begin{remark}
 The proposed approach allows also to address the situation in which regressors are not a.s. bounded. For instance, consider the case of random regressors with i.i.d sub-Gaussian entries such that
\[
\forall j\leq n,\quad \bE\left[\exp\left(\tfrac{[\phi_i]_{j}^2}{\varkappa^2}\right)\right]\leq 1.
\] 
Using the fact that the maximum of uniform norms $\|\phi_i\|_{\infty}$, $1\leq i\leq m$, concentrates around $\varkappa\sqrt{\ln mn}$ along with independence of noises $\xi_i$  of $\phi_i$, the ``smoothness'' and ``sub-Gaussianity'' assumptions of Proposition \ref{cor:mycorL1} can be stated ``conditionally'' to the event $\left\{\omega:\;\max_{i\leq m}\|\phi_i\|_{\infty}^2\lesssim \varkappa^2(\ln [mn]+t)\right\}$ of probability greater than $1-e^{-t}$. For instance, when replacing the bound for the uniform norm of regressors with $\varkappa^2(\ln [mn]+t)$ in the definition of algorithm parameters and combining with appropriate deviation inequality for martingales (cf., e.g., \cite{bercu2015concentration}), one arrives at the bound for the error $\|\widehat{x}_N - x_*\|_1$ of Algorithm \ref{alg2} which is similar to \rf{eq:prop32nob} of Proposition \ref{cor:mycorL1} in which $\ov\nu$ is replaced
with $\varkappa\sqrt{\ln [mn]+t}$.
\end{remark}
\subsection{Numerical experiments}\label{sec:num}
\begin{figure}[H]
\centering
\begin{center}
\begin{tabular}{cc}
\includegraphics[width=0.5\textwidth]{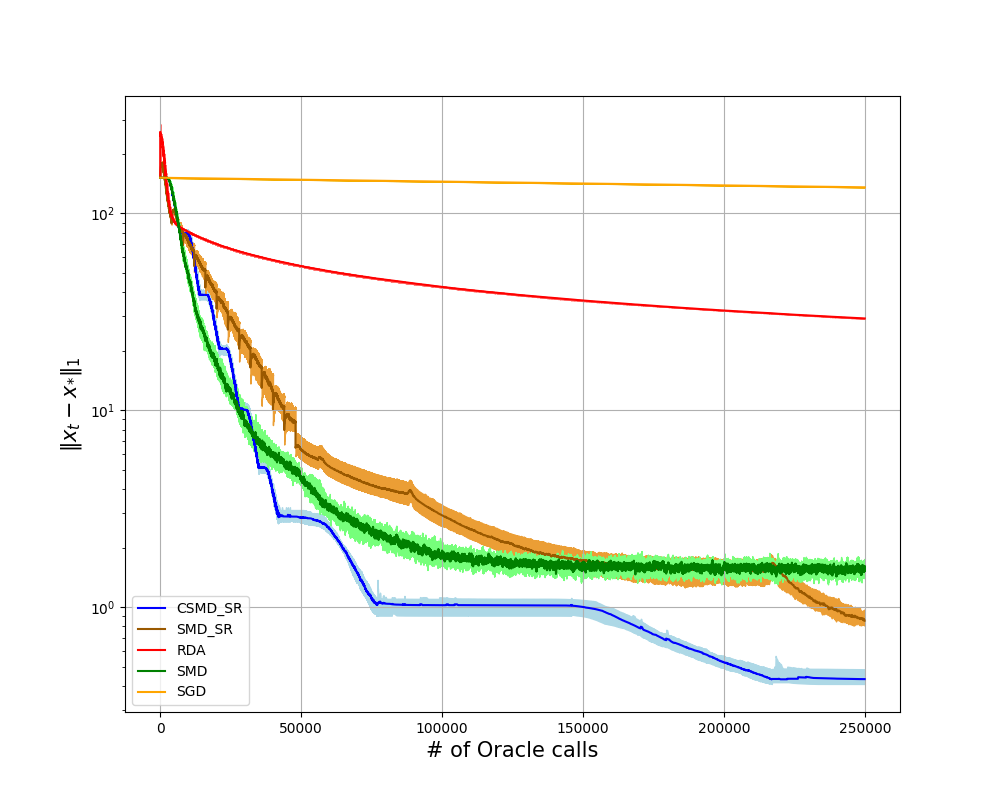}
&\includegraphics[width=0.5\textwidth]{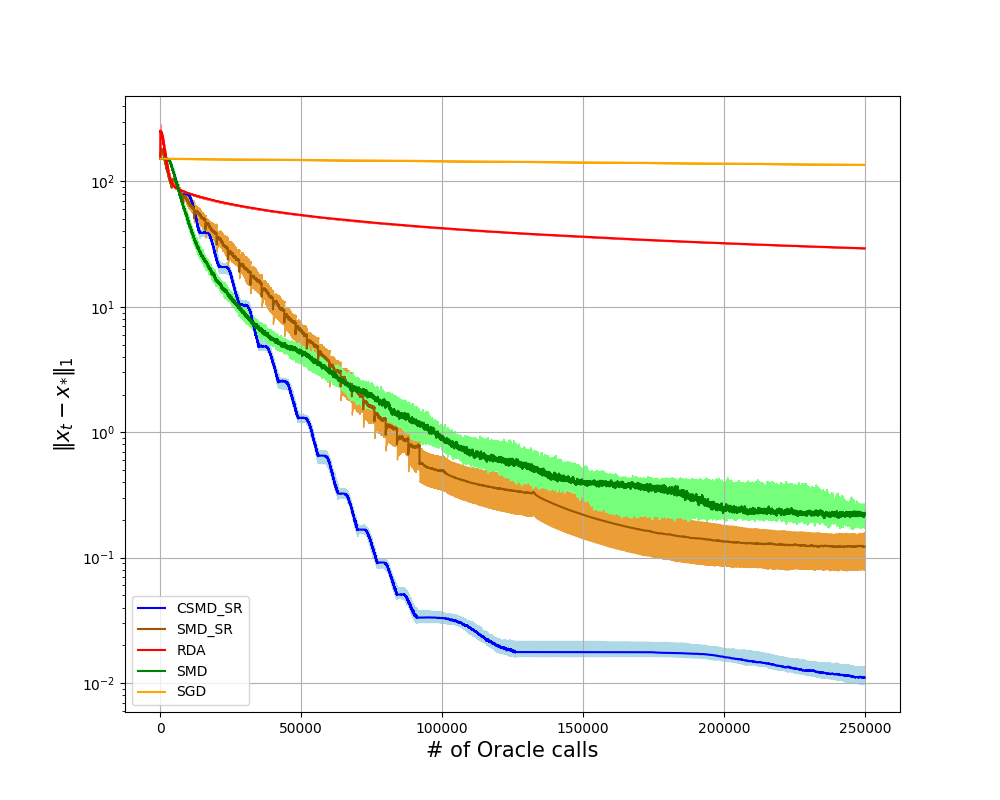}\\
 $\sigma = 0.1 $&$\sigma = 0.001 $
\end{tabular}
\end{center}
\caption{\small CSMD-SR and ``vanilla'' SMD in Generalized Linear Regression problem: $\ell_1$ error as a function of the number of oracle calls\small\label{fig:fig2}}
\end{figure}

In this section, we present results of a small simulation study illustrating the theoretical part of the previous section.\footnote{The reader is invited to check Section \ref{sec:suppexp} of the supplementary material for more experimental results.} We consider the GLR model \rf{sparselin} with activation function \eqref{ral} where $\alpha=1/2$.
In our simulations, $x_*$ is an $s$-sparse vector with $s$ nonvanishing components sampled independently from the standard $s$-dimensional Gaussian distribution; regressors $\phi_i$ are sampled from a multivariate Gaussian distribution $\phi \sim \mathcal{N}(0,\Sigma)$, where $\Sigma$ is a diagonal covariance matrix with diagonal entries $\sigma_{1} \leq ...\leq \sigma_{n}$. In Figure \ref{fig:fig2} we report on the experiment in which we compare the performance of the CSMD-SR algorithm from Section \ref{sec:multi_alg} to that of four other methods. The contenders are (1) ``vanilla'' non-Euclidean SMD algorithm constrained to the $\ell_1$-ball equipped with the distance generating function \eqref{proxfct}, (2) composite non-Euclidean dual averaging algorithm ($p$-Norm RDA) from \cite{JMLR:v11:xiao10a}, (3) multistage SMD-SR of \cite{juditsky2020sparse}, and (4) ``vanilla'' Euclidean SGD.
The  regularization parameter of the $\ell_1$ penalty in (2) is set to the theoretically optimal value $\lambda = 2\sigma\sqrt{2 \log(n) / T}$. The corresponding dimension of the parameter space is $n=500000$, the sparsity level  of the optimal point $x_*$ is $s = 200$, and the ``total budget'' of oracle calls is  $ N=250000$; we use the identity regressor covariance matrix ($\Sigma=I_n$) and $\sigma\in\{0.001, 0.1\}$.
To reduce computation time we use the minibatch versions of the multi-stage algorithms---CSMD-SR and algorithm (3)), the data to compute stochastic gradient realizations $\nabla G(x_i, \omega)=\phi (\myu(\phi^Tx_i)-\eta)$ at the current search point $x_i$ being generated ``on the fly.''
We repeat simulations 20 times and plot the median value along with the first and the last deciles of the error
$\|\wh x_i-x_*\|_1$  at each iteration of the algorithm against the number of oracle calls.

The proposed method outperforms other algorithms which struggle to reach the regime where the stochastic noise is dominant. 

\begin{figure}[H]
    \centering
  \begin{subfigure}[b]{0.5\linewidth}
    \centering
    \includegraphics[width=1.\linewidth]{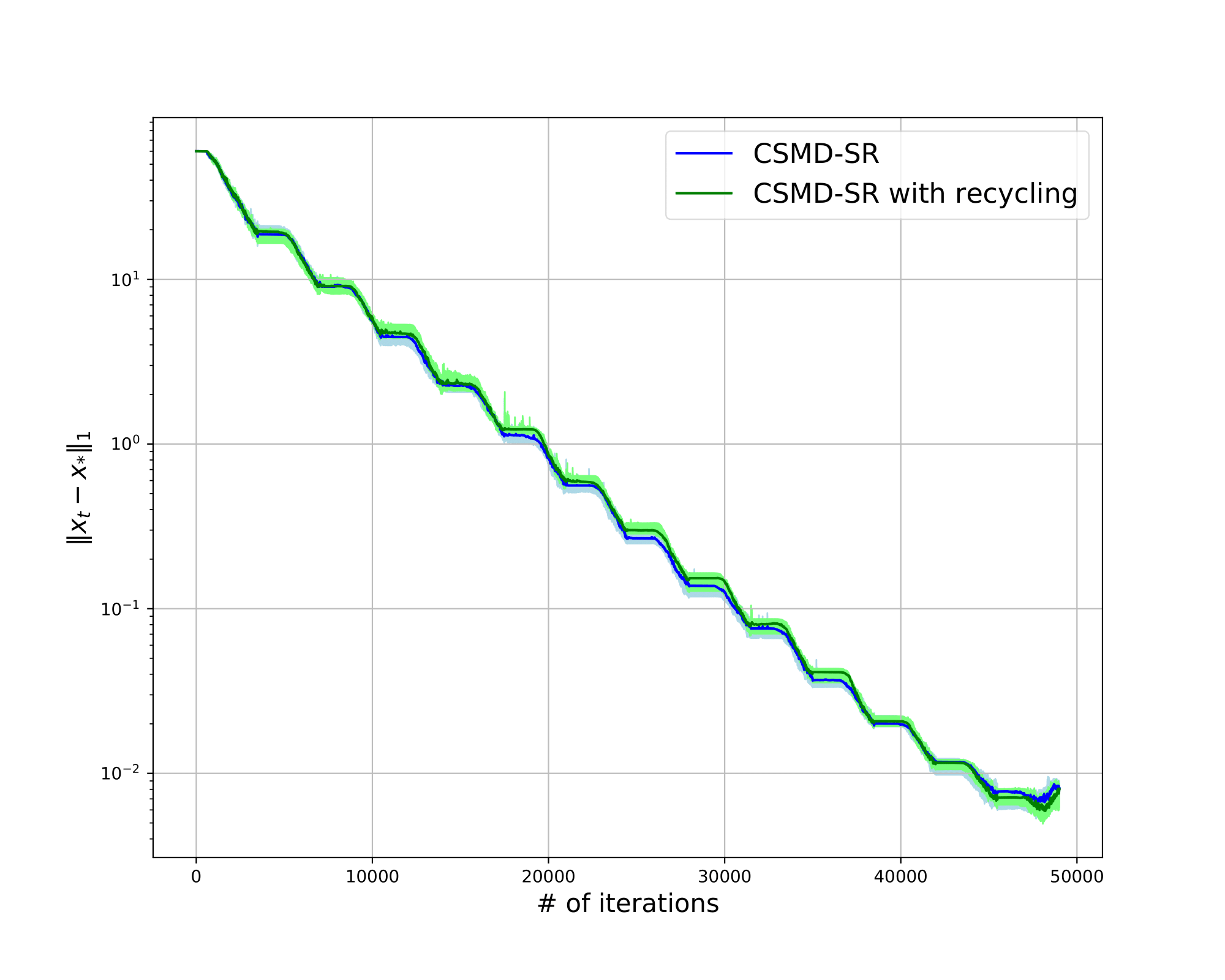}
    \label{fig7:a}
    \vspace{3ex}
  \end{subfigure}
  \begin{subfigure}[b]{0.5\linewidth}
    \centering
    \includegraphics[width=1.\linewidth]{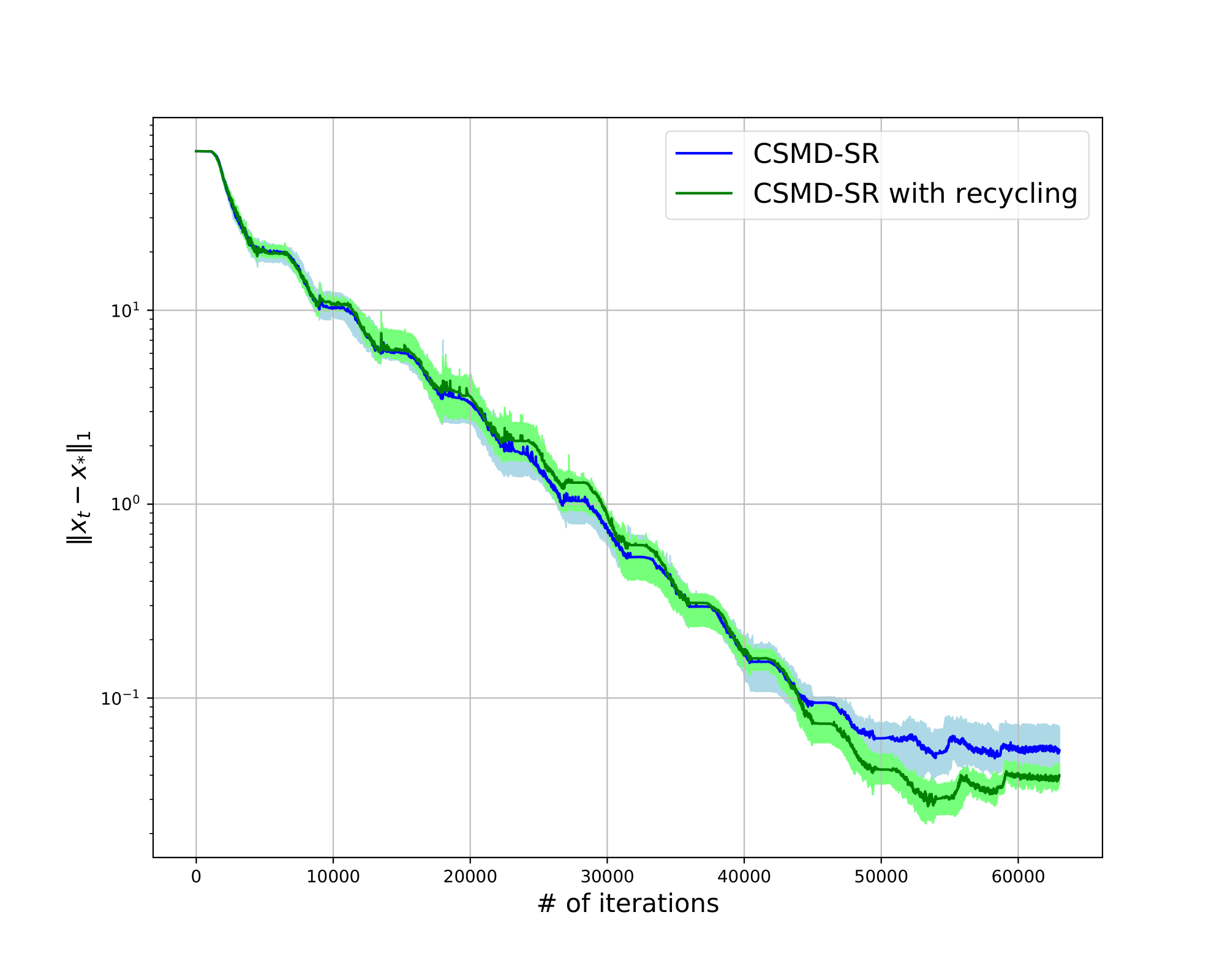}
    \label{fig7:b} 
    \vspace{3ex}
  \end{subfigure}\caption{\small \label{fig6} Preliminary stages of the CSMD-SR and its variant with data recycling: linear regression experiment (left pane), GLR with activation   $\myu_{1/10}(t)$ (right pane).}
\end{figure}

In the second experiment we report on here, we study the behavior of the multistage algorithm derived from Algorithm \ref{alg2-minibatch} in which, instead of using independent data samples, we reuse the same data at each stage of the method. In Figure \ref{fig6} we present results of comparison of the CSMD-SR algorithm with its variant with data recycle. This version is of interest as it attains fast the noise regime while using  limited amount of samples.
In our first experiment, we consider linear regression problem with parameter dimension $n = 100\,000$ and sparsity level $s = 75$ of the optimal solution; we consider the GLR model \eqref{sparselin} with activation function $\myu_{1/10}(t)$ in the second experiment. We choose $\Sigma = I_n$ and $\sigma = 0.001$; we run 14 (preliminary) stages of the algorithm with $m_0= 3500$ in the first simulation and $m_0 = 4500$ in the second. We believe that the results speak for themselves.

\vspace{0.5cm}
\noindent{\Large\textbf{Acknowledgements}}\\\\
This work was supported by Multidisciplinary Institute in Artificial intelligence MIAI {@} Grenoble Alpes (ANR-19-P3IA-0003),  ``Investissements d’avenir'' program (ANR20-CE23-0007-01), FAIRPLAY project, LabEx Ecodec (ANR11-LABX-0047), and ANR-19-CE23-0026. The authors would also like to acknowledge CRITEO AI Lab for supporting this work.


\newpage
\appendix

\newpage\section{Proofs}
We use notation $\bE_{i}$ for conditional expectation given $x_0$ and $\omega_1,...,\omega_{i}$.
\subsection{Proof of Proposition \ref{pr:mupropn}}
The result of Proposition \ref{pr:mupropn} is an immediate consequence of the following statement.
\begin{proposition}\label{pr:muprop}
Let
\[f(x)=\half g(x)+h(x), \quad x\in X.
\]
In the situation of { Section \ref{sec:CSMD}}, let $\gamma_i\le (4\nu)^{-1}$ for all $i=0,1,...$, and let  $\widehat{x}_m$ be defined in \rf{eq:asol}, where $x_i$
		are iterations \rf{eq:fa}. Then for any $t\geq 2\sqrt{2+\ln m}$ there is $\overline \Omega_m\subset \Omega$ such that $\Prob(\overline \Omega_m)\geq 1-4e^{-t}$ and  for all $\omega^m=[\omega_1,...,\omega_m]\in   \overline \Omega_m$,
		\begin{align}
		\left(\sum_{i=0}^{m-1}\gamma_i\right)[f(\widehat{x}_m)-f(x_*)]&\le\sum_{i=0}^{m-1}\Big[\half \gamma_{i}\la \nabla g(x_{i}), x_{i}-x_* \ra+\gamma_{i+1} (h(x_{i+1})-h(x_*))\Big]
		\nn
		&\leq  V({x_0},x_*)+\gamma_0[h(x_0)-h(x_*)]-\gamma_m[h(x_m)-h(x_*)]\nn
&\qquad+V(x_{0},x_*)+{15}t R^2+\sigma^2_*\left[{7}\sum_{i=0}^{m-1}\gamma_i^2+24t\overline \gamma^2\right].
		\label{simple1}
\end{align}
In particular, when using the constant stepsize strategy with $\gamma_i\equiv \gamma$,  $0<\gamma\le (4\nu)^{-1}$, one has
\begin{align}
&\half [g(\wh x_{m})-g(x_*)]+[h(\wh x_m)-h(x_*)]
\nn&\qquad\leq {V(x_{0},x_*)+{15}t R^2\over \gamma m}+{h(x_0)-h(x_m)\over m}+\gamma\sigma^2_*\left({7}+{24t\over m}\right).
\label{eq:almo_a}
\end{align}
	\end{proposition}
\paragraph{Proof.}
Denote $H_i=\nabla G(x_{i-1},\omega_i)$. In the sequel, we use the shortcut notation $\vartheta(z)$ and $V(x,z)$ for  $\vartheta_{x_0}^R(z)$ and $V_{x_0}(x,z)$ when exact values $x_0$ and $R$ are clear from the context.
\paragraph{1$^o$.}  From the  definition of $x_i$ and  of the composite prox-mapping \rf{cprox}  (cf. Lemma A.1 of \cite{nesterov2013first}), we conclude that there is  $\eta_i\in \partial h(x_i)$ such that
\[
	\la \gamma_{i-1} H_i+\gamma_i\eta_i+\nabla\vartheta(x_i)-\nabla\vartheta(x_{i-1}),z-x_{i}\ra\ge 0,\;\;\forall \; z\in \X,
\]
implying, as usual \cite{chen1993convergence}, that $\forall z\in \X$
\[
\la \gamma_{i-1}  H_i+\gamma_i\eta_i, x_{i}-z \ra\leq V(x_{i-1},z)-V(x_{i},z)-V(x_{i-1},x_{i}).
\]
In particular,
\begin{align*}
&\gamma_{i-1} \la H_i, x_{i-1}-x_* \ra+\gamma_i \la\eta_i, x_{i}-x_* \ra\\&\leq V(x_{i-1},x_*)-V(x_{i},x_*)-V(x_{i-1},x_{i})
+\gamma_{i-1} \la H_i, x_{i-1}-x_i \ra\\&\leq V(x_{i-1},x_*)-V(x_{i},x_*)
+\half\gamma^2_{i-1} \| H_i\|_*^2.
\end{align*}
Observe that due to the Lipschitz continuity of $ \nabla G(\cdot,\omega)$ one has
\be
\nu\la  \nabla G(x,\omega)- \nabla G(x',\omega),x-x'\ra \geq \| \nabla G(x,\omega)- \nabla G(x',\omega)\|_*^2,\quad \forall x,x'\in \X,
\ee{eq:dstrong}
so that
\begin{align*}
\| \nabla G(x,\omega)\|_*^2&\leq 2\| \nabla G(x,\omega)- \nabla G(x_*,\omega)\|_*^2+2\| \nabla G(x_*,\omega)\|_*^2\nn
&\leq 2\nu\la  \nabla G(x,\omega)- \nabla G(x_*,\omega),x-x_*\ra +2\| \nabla G(x_*,\omega)\|_*^2\nn
&=2\nu\la  \nabla G(x,\omega),x-x_*\ra-2\nu\la  \nabla G(x_*,\omega),x-x_*\ra+2\| \nabla G(x_*,\omega)\|_*^2
\end{align*}so that
\begin{align*}
&\gamma_{i-1} \la H_i, x_{i-1}-x_* \ra+\gamma_i \la\eta_i,, x_{i}-x_* \ra
\\&\leq V(x_{i-1},x_*)-V(x_{i},x_*)
+\gamma^2_{i-1}[\nu \la H_i, x_{i-1}-x_* \ra- \nu\zeta_i+\tau_i]
\end{align*}
where $\zeta_i=\la  \nabla G(x_*,\omega_i),x_{i-1}-x_*\ra$ and $\tau_i=\| \nabla G(x_*,\omega)\|_*^2$.
As a result, by convexity of  $h$ we have for $\gamma_i\leq (4\nu)^{-1}$
\begin{align*}
&\tfo\gamma_{i-1}\la \nabla g(x_{i-1}), x_{i-1}-x_* \ra+\gamma_i[h(x_i)-h(x_*)]\\
&\leq (\gamma_{i-1}-\gamma_{i-1}^2\nu)\la \nabla g(x_{i-1}), x_{i-1}-x_* \ra+\gamma_i \la\eta_i, x_{i}-x_* \ra\\
&\leq V(x_{i-1},x_*)-V(x_{i},x_*)
+ (\gamma_{i-1}-\gamma_{i-1}^2\nu)\la \xi_{i},x_{i-1}-x_*\ra +\gamma_{i-1}^2[\tau_i-\nu\zeta_i]
\end{align*}
where we put $\xi_i= H_i-\nabla g(x_{i-1})$.
When summing from $i=1$ to $m$ we obtain
\begin{align}
&\sum_{i=1}^m\gamma_{i-1}\Big(\tfo\la \nabla g(x_{i-1}), x_{i-1}-x_* \ra+[h(x_{i-1})-h(x_*)]\Big)\nonumber\\
&\qquad\leq V(x_{0},x_*)+\underbrace{\sum_{i=1}^{m}[\gamma_{i-1}^2(\tau_i-\nu\zeta_i)+\gamma_{i-1}(1-\gamma_{i-1}\nu)\la \xi_{i},x_{i-1}-x_*\ra]}_{=:R_m}\nonumber\\
&\qquad\qquad+\gamma_0[h(x_0)-h(x_*)]-\gamma_m[h(x_m)-h(x_*)]. 
\label{eq:almo_2}
\end{align}
\paragraph{2$^o$.} We have
\bse
\gamma_{i-1}\la \xi_{i},x_{i-1}-x_*\ra&=&\gamma_{i-1}\overbrace{\la [\nabla G(x_{i-1},\omega_{i})-\nabla G(x_*,\omega_{i})]-\nabla g(x_{i-1}),x_{i-1}-x_*\ra}^{\upsilon_i }\\&&+\gamma_{i-1}\la \nabla G(x_*,\omega_{i}),x_{i-1}-x_*\ra\\
&=&\gamma_{i-1}[\upsilon_i+\zeta_i],
\ese
so that
\be
R_m={\sum_{i=1}^{m} \gamma_{i-1}^2\tau_i}+{\sum_{i=1}^{m}(\gamma_{i-1}-\gamma^2_{i-1}\nu)\upsilon_i}
+{\sum_{i=1}^{m}(\gamma_{i-1}-2\nu\gamma_{i-1}^2)\zeta_i}=:r^{(1)}_m+r^{(2)}_m+r^{(3)}_m.
\ee{eq:RT}
Note that $r^{(3)}_m$ is a sub-Gaussian martingale. Indeed, one has $\bE_{i-1}\{\zeta_i\}=0$ a.s.,\footnote{We use notation $\bE_{i-1}$ for the conditional expectation given $x_0,\omega_1,...,\omega_{i-1}$.} and
\[|\zeta_i|\leq \|x_{i-1}-x_*\|\,\|\nabla G(x_*,\omega)\|_*,\]
so that by the sub-Gaussian hypothesis \rf{eq:subgaus}, $\bE_{i-1}\Big\{\exp\Big(\underbrace{\zeta^2_i\over 4R^2\sigma_*^2}_{\nu_*^2}\Big)\Big\}\leq \exp(1)$.
As a result (cf. the proof of Proposition 4.2 in \cite{juditsky2008large}),
\[
\forall t\quad\bE_{i-1}\left\{e^{t\zeta_i}\right\}\leq \exp\left(t\bE_{i-1}\{\zeta_i\}+{\tfrac{3}{4}t^2\nu^2_*}\right)=\exp\left({3t^2R^2\sigma^2_*}\right),
\]
and applying \rf{eq:subg1} to $S_m=r_m^{(3)}$ with \[
r_m=6R^2\sigma_*^2\sum_{i=0}^{m-1}(\gamma_{i}-2\nu\gamma_i^2)^2\leq{6}R^2\sigma_*^2\sum_{i=0}^{m-1}\gamma^2_{i}\] we conclude that for some $\Omega^{(3)}_m$ such that $\Prob(\Omega^{(3)}_m)\geq 1-e^{-t}$ and all $\omega^m\in \Omega^{(3)}_m$
\be
r^{(3)}_m\leq2\sqrt{{3t R^2\sigma_*^2}\sum_{i=0}^{m-1}\gamma_{i}^2}\leq {{3}tR^2}+{3}\sigma_*^2\sum_{i=0}^{m-1}\gamma_{i}^2.
\ee{r3}
Next, again by \rf{eq:subgaus}, due to the Jensen inequality, $\bE_{i-1}\{\tau_i\}\leq \sigma_*^2$, and
\[
\bE_{i-1}\left\{\exp\left(t\|\nabla G(x_*,\omega_i)\|_*\right)\right\}\leq \exp\left(t\bE_{i-1}\{\|\nabla G(x_*,\omega_i)\|_*\}+{\tfrac{3}{4}t^2\sigma^2_*}\right)\leq\exp\left(t\sigma_*
+\tfrac{3}{4}t^2\sigma^2_*\right).
\]
Thus, when setting
\[
\mu_i=\gamma_{i-1}\sigma_*,\;\;s_i^2=\tfrac{3}{2}\gamma_{i-1}\sigma_*^2,\;\;\overline s=\max_i \gamma_is_i,
\]
$M_m=r^{(1)}_m$,
$v_m+h_m=\tfrac{21}{4}\sigma_*^4\sum_{i=0}^{m-1}\gamma_i^4$,
and applying the bound \rf{eq:subg2} of Lemma \ref{lem:mart1} we obtain
\[
r^{(1)}_m\leq 3\sigma^2_*\sum_{i=0}^{m-1}\gamma_i^2+\underbrace{\sqrt{21t \sigma_*^4\sum_{i=0}^{m-1}\gamma_{i}^4}}_{=:\Delta^{(1)}_m}+3t\overline \gamma^2\sigma_*^2
\]
 for $\overline \gamma=\max_i\gamma_i$ and $\omega^m\in \Omega^{(1)}_m$ where $\Omega^{(1)}_m$ is of probability at least $1-e^{-x}$. Because
 \[
 \overline \gamma^2\sum_{i=0}^{m-1}\gamma_{i}^2\geq \sum_{i=0}^{m-1}\gamma_{i}^4,
 \]
whenever
 $
 \sqrt{21t \sigma_*^4\sum_{i=0}^{m-1}\gamma_{i}^4}\geq \sum_{i=0}^{m-1}\gamma_i^2,
 $ one has
 $
 21t\overline \gamma^2\geq \sum_{i=0}^{m-1}\gamma_{i}^2
 $ and
 \[21t \sum_{i=0}^{m-1}\gamma_{i}^4\leq {21t\overline \gamma^2\sum_{i=0}^{m-1}\gamma_{i}^2}\leq (21t\overline \gamma^2)^2
 \]

 Thus,
 \[
 \Delta^{(1)}_m\leq \min\left[21t\sigma^2_*\overline \gamma^2,\sigma_*^2\sum_{i=0}^{m-1}\gamma_i^2\right]\leq 21t\sigma^2_*\overline \gamma^2+\sigma_*^2\sum_{i=0}^{m-1}\gamma_i^2,
 \]
 and
 \be
 r^{(1)}_m\leq \sigma^2_*\left[4\sum_{i=0}^{m-1}\gamma_i^2+24t\overline \gamma^2\right]
 \ee{r1}
 for $\omega^m\in \Omega^{(1)}_m$.
 \par
Finally, by the Lipschitz continuity of $\nabla G$ (cf. \rf{eq:dstrong}), when taking expectation w.r.t. the distribution of $\omega_i$, we get
\bse
\bE_{i-1}\{\upsilon_i^2\}&\leq& 4R^2\bE_{i-1}\{\|\nabla G(x_{i-1},\omega_{i})-\nabla G(x_*,\omega_{i})\|_*^2\}\\&\leq& 4R^2\nu\bE_{i-1}\{\la \nabla G(x_{i-1},\omega_{i})-\nabla G(x_*,\omega_{i}),x_{i-1}-x_*\ra\}
=4R^2\nu\la \nabla g(x_{i-1}),x_{i-1}-x_*\ra.
\ese
On the other hand, one also has $|\upsilon_i|\leq 2\nu\|x_{i-1}-x_i\|^2\leq 8\nu R^2$. We can now apply Lemma \ref{lem:mart} with
$\sigma_i^2=4\gamma_{i-1}^2 R^2\nu\la \nabla g(x_{i-1}),x_{i-1}-x_*\ra$ to conclude that for $t\geq  2\sqrt{2+\ln m}$
\[r_m^{(2)}\leq {4}\underbrace{\sqrt{tR^2\nu\sum_{i=0}^{m-1} \gamma_i^2 \la \nabla g(x_{i}),x_{i}-x_*\ra}}_{=:\Delta^{(2)}_m}+{{16}t\nu R^2\overline\gamma}
\]
for all $\omega^m\in \Omega^{(2)}_m$ such that $\Prob(\Omega^{(2)}_m)\geq 1-2e^{-t}$. Note that
\[\Delta_m^{(2)}\leq {2}{tR^2}+{\tfrac{1}{4}}\nu\sum_{i=0}^{m-1}\gamma_i^2 \la \nabla g(x_{i}),x_{i}-x_*\ra,
\]
and $\gamma_i\leq (4\nu)^{-1}$, so that
\be
r_m^{(2)}\leq \nu\sum_{i=0}^{m-1} \gamma_i^2 \la \nabla g(x_{i}),x_{i}-x_*\ra+{12}{t R^2}\leq \tfrac{1}{4} \sum_{i=0}^{m-1} \gamma_i \la \nabla g(x_{i}),x_{i}-x_*\ra+{12}{t R^2}
\ee{r2}
for $\omega^m\in \Omega^{(2)}_m$.
\paragraph{3$^o$.}
When substituting bounds \rf{r3}--\rf{r2} into \rf{eq:RT} we obtain
\bse
R_m&\leq& \tfrac{1}{4}\sum_{i=0}^{m-1} \gamma_i \la \nabla g(x_{i}),x_{i}-x_*\ra+{12}{t R^2}+\sigma^2_*\left[4\sum_{i=0}^{m-1}\gamma_i^2+24t\overline \gamma^2\right]+{2}\sqrt{{3}t R^2\sigma_*^2\sum_{i=0}^{m-1}\gamma_{i}^2}\\
&\leq&\tfrac{1}{4}\sum_{i=0}^{m-1} \gamma_i \la \nabla g(x_{i}),x_{i}-x_*\ra+{15}t R^2+\sigma^2_*\left[{7}\sum_{i=0}^{m-1}\gamma_i^2+24t\overline \gamma^2\right]
\ese
for all $\omega^m\in \overline\Omega_m=\bigcap_{i=1}^3\Omega^{(i)}_m$ with
$\Prob (\overline\Omega_m)\geq 1-4e^{-t}$ and $t\geq  2\sqrt{2+\ln m}$.

When substituting the latter bound into \rf{eq:almo_2} and utilizing the convexity of $g$ and $h$ we arrive at
\begin{align*}
&\left(\sum_{i=0}^{m-1}\gamma_i\right)\Big(\half [g(\wh x_{m})-g(x_*)]+[h(\wh x_m)-h(x_*)]\Big)\leq \sum_{i=0}^{m-1}\gamma_{i}\Big(\half[g(x_{i})- g(x_*)]+[h(x_{i})-h(x_*)]\Big)\\
&\leq \sum_{i=1}^m\gamma_{i-1}\Big(\half\la \nabla g(x_{i-1}, x_{i-1}-x_* \ra+[h(x_{i-1})-h(x_*)]\Big)\\
&\leq V(x_{0},x_*)+{15}t R^2+\sigma^2_*\left[{7}\sum_{i=0}^{m-1}\gamma_i^2+24t\overline \gamma^2\right]+\gamma_0[h(x_0)-h(x_*)]-\gamma_m[h(x_m)-h(x_*)]. 
\end{align*}
In particular, for constant stepsizes $\gamma_i\equiv\gamma$ we get
\begin{align*}
&\half [g(\wh x_{m})-g(x_*)]+[h(\wh x_m)-h(x_*)]
\\&\quad\leq {V(x_{0},x_*)+{15}t R^2\over \gamma m}+{h(x_0)-h(x_m)\over m}+\gamma\sigma^2_*\left({7}+\frac{24t}{ m}\right). 
\end{align*}
This implies the first statement of the proposition.
\paragraph{5$^o$.} To prove the bound for the minibatch solution $\wh x_m^{(L)} =\left(\sum_{i=0}^{m-1}\gamma_i\right)^{-1}\sum_{i=0}^{m-1} \gamma_ix_i^{(L)}$,
it suffices to note that minibatch gradient observation $H(x,\omega^{(L)})$ is Lipschitz-continuous with Lipschitz constant $\nu$, and that
$H(x_*,\omega^{(L)})$ is sub-Gaussian with parameter $\sigma_*^2$  replaced with   $\ov\sigma^2_{*,L}\lesssim{\Theta\sigma_*^2\over L}$, see Lemma \ref{lem:sum}.\qed

\subsection{Deviation inequalities}
Let us assume that  $(\xi_i, \F_i)_{i = 1,2,...}$ is a sequence of sub-Gaussian random variables satisfying\footnote{Here, same as above, we denote $\bE_{i-1}$ the expectation conditional to $\F_{i-1}$.}
\begin{equation}
    \bE_{i-1}\left\{ e^{t\xi_i}\right\} \leq e^{t\mu_i +\frac{t^2 s_i^2}{2} }, \hspace{0.5cm} a.s. \label{subgauss}
\end{equation}
for some {\em nonrandom} $\mu_i, s_i$, $s_i\leq \overline s$.
We denote by $S_n = \sum_{i=1}^n{\xi_i - \mu_i}$, $r_n= \sum_{i=1}^n{s_i^2}$, $v_n =\sum_{i=1}^n{s_i^4}, M_n = \sum_{i=1}^n \xi_i^2 - (s_i^2 + \mu_i^2)$, and $h_n = \sum_{i=1}^n 2\mu_i^2 s_i^2.$ The following well known result is provided for reader's convenience.
\begin{lemma}
\label{lem:mart1}
For all $x>0$ one has
\begin{subequations}
\begin{align}
\label{eq:subg1}
\Prob\left\{S_n\geq \sqrt{2xr_n}\right\}&\leq e^{-x},\\
\Prob\left\{M_n\geq 2\sqrt{x(v_n+h_n)}+2x{\overline s}^2
\right\}&\leq e^{-x}.
\label{eq:subg2}
\end{align}
\end{subequations}
\end{lemma}
\paragraph{Proof.}
The inequality \rf{eq:subg1} is straightforward. To prove \rf{eq:subg2}, note that for $t <\frac{1}{2}\overline{s}^{-2}$ and $\eta \sim \mathcal{N}(0,1)$ independent of $\xi_0, ..., \xi_n$ , we have:
\begin{align*}
    \bE_{i-1}\left\{ e^{t\xi_i^2}\right\} &= \bE_{i-1}\left\{ \bE_{\eta}\left\{ e^{\sqrt{2t}\xi_i \eta}\right\}\right\} = \bE_{\eta}\left\{ \bE_{i-1}\left\{ e^{\sqrt{2t}\xi_i \eta}\right\}\right\} \\
    & \le \bE_{\eta}\left\{ \exp\left\{\sqrt{2t}\eta \mu_i + t\eta^2 s_i^2\right\}\right\} = (1 - 2ts_i^2)^{-1/2}\exp\left\{\frac{t\mu_i^2}{1-2ts_i^2}\right\} \hspace{0.2cm}\text{a.s.},
\end{align*}
and because, cf \cite[Lemma 1]{laurent2000adaptive},
\begin{align*}
    -\tfrac{1}{2}\text{ln}(1-2ts_i^2) + \frac{t\mu_i^2}{1-2ts_i^2} - t(s_i^2 +\mu_i^2) \leq \frac{t^2 s_i^2(s_i^2 +2\mu_i^2)}{1-2ts_i^2} \leq \frac{t^2 s_i^2(s_i^2 +2\mu_i^2)}{1-2t\overline{s}^2},
\end{align*}
one has
for $t < \tfrac{1}{2}\overline{s}^{-2}$
\[
    \bE\left\{ e^{tM_n}
    \right\} \leq \exp\left\{ \frac{t^2 (v_n +h_n)}{1-2t\overline{s}^2}\right\}.
\] By Lemma 8 of \cite{birge1998minimum}, this implies that
\[
\Prob\left\{M_n\geq 2\sqrt{x(v_n+h_n)}+2x{\overline s}^2
\right\}\leq e^{-x}
\]
for all $x>0$.\qed

Now, suppose that $\zeta_i,\;i=1,2,...$ is a sequence of random variables satisfying
\be
\bE_{i-1}\{\zeta_i\}=\mu_i,\,\bE_{i-1}\{\zeta_i^2\}\leq \sigma_i^2,\;\;|\zeta_i|\leq 1\;\;\mathrm{a.s.}
\ee{subg}
Denote $M_n=\sum_{i=1}^n [\zeta_i-\mu_i]$ and  $q_n=\sum_{i=1}^n \sigma_i^2$. Note that $q_n\leq n$.
\begin{lemma}\label{lem:mart}
Let $x\geq 1$; one has
\[
\Prob\left\{M_n\geq \sqrt{2xq_n}+{x}\right\}\leq \left[e\left(2x\ln\left[{9n\over 2x}\right]+1\right)+1\right]e^{-x}.
\]
In particular, for $x\geq 4\sqrt{2+\ln n}$ one has
\[
\Prob\left\{M_n\geq \sqrt{2xq_n}+{x}\right\}\leq 2e^{-x/2}.
\]
\end{lemma}
{\bf Proof.}
In the premise of the lemma, applying Bernstein's inequality for martingales \cite{bercu2015concentration,fan2012hoeffding}  we obtain for all $x>0$ and $u>0$,
\[
\Prob\left\{M_n\geq \sqrt{2xu}+{x\over 3},\,q_n\leq u\right\}\leq e^{-x}.
\]
We conclude that
\[
\Prob\left\{M_n\geq {x},\,q_n\leq {2x\over 9}\right\}\leq e^{-x},
\]
and for any $u>0$
\[
\Prob\left\{M_n\geq \sqrt{2(x+1)q_n}+{x\over 3},\,u\leq q_n\leq u\big(1+{1/x}\big)\right\}\leq e^{-x},
\]
so that
\[
\delta_n(x;u):=\Prob\left\{M_n\geq \sqrt{2xq_n}+{x\over 3},\,u\leq q_n\leq u\big(1+{1/x}\big)\right\}\leq e^{-x+1}.
\]
Let now $u_0=2x/ 9$, $u_j=\min\{n,(1+1/x)^j u_0\}$, $j=0,..., J$, with
\[
J=\left\rfloor\ln\big[{n/ u_0}\big]\ln^{-1}[1+1/x]\right\lfloor.
\]
Note that $\ln[1+1/x]\geq 1/(2x)$ for $x\geq 1$, so that
\[J \leq \ln\big[{n/u_0}\big]\ln^{-1} [1+1/x]+1\leq 2x\ln\big[{n/u_0}\big]+1.
\]
On the other hand,
\bse
\Prob\left\{M_n\geq \sqrt{2xq_n}+{x}\right\}&\leq&e^{-x}+\sum_{j=1}^{J}  \delta_n(x;u_j)\leq e^{-x}+J e^{-x+1}\\&\leq &\Big[e\Big(2x\ln\Big[{9n\over 2x}\Big]+1\Big)+1\Big]e^{-x}
\ese
Finally, we verify explicitly that for $x\geq 4\sqrt{2+\ln n}$ one has
\[
\Big[e\Big(2x\ln\Big[{9n\over 2x}\Big]+1\Big)+1\Big]e^{-x/2}\leq 2,
\]
implying that for such $x$
\[
\Prob\left\{M_n\geq \sqrt{2xq_n}+{x}\right\}\leq 2e^{-x/2}.\eqno{\mbox{\qed}}
\]
Let $(\xi_i)_{i=1,...}$ be a sequence of independent random vectors in $\bR^n$ such that
\[
\bE_{i-1}\left\{\exp\left({\|\xi_i\|_*^2\over s^2}\right)\right\}\leq \exp(1),
\]
and let $\eta=\sum_{i=1}^m\xi_i$, $m\in \bZ_+$. We are interested in ``sub-Gaussian characteristics'' of r.v.
$\zeta=\la u,\eta\ra$ for some $u\in \bR^n$, $\|u\|\leq R$, and of $\tau= \|\eta\|_*$.
\par
Because $\bE\{\la u,\xi_i\ra\}=0$ and $|\la u,\xi_i\ra|\leq \|u\|\,\|\xi_i\|_*$, for all $t$ one has (cf.,e.g., Proposition 4.2 of \cite{juditsky2008large})
\[
\bE\Big\{e^{t\la u,\eta\ra}\Big\}=\prod_{i=1}^m \bE\Big\{e^{t\la u,\xi_i\ra}\Big\}\leq \prod_{i=1}^m \exp\left({\tfrac{3}{4}t^2s^2}\right)=\exp\left(\tfrac{3}{4}mt^2s^2\right).
\]
Let $\xi_\ell$, $\ell=1,2,...$ be a sequence of independent random vectors $\xi_\ell\in E$, such that $\bE\{\xi_\ell\}=0$ and $\bE\Big\{e^{
\|\xi_\ell\|_*^2/s^2}\Big\}\leq \exp(1)$. Denote $\eta_j=\sum_{\ell=1}^j\xi_\ell$.
We have the following result.
\begin{lemma}\label{lem:sum}
\be\forall L \in \bZ_+\quad
\bE\left\{\exp\left({\|\eta_L \|_*^2 \over 10\Theta s^2 L }\right)\right\}\leq \exp(1)
\ee{sssg}
where $\Theta=\max_{\|z\|\leq 1}\theta(z)$ for the d.-g.f. $\theta$ of the unit ball of norm $\|\cdot\|$ in $E$, as defined in Section \ref{sec:CSMD}.
\end{lemma}
{\bf Proof.}
Let for $\eta\in E$, $\pi(\eta)=\sup_{\|z\|\leq 1}[\la \eta,z\ra-\theta(z)]$.
Observe that for all $\beta>0$,
\be\|\eta_L \|_*=\sup_{\|z\|\leq 1}\la \eta_L ,z\ra\leq
\max_{\|z\|\leq 1}\beta\theta(z)+\beta \pi(\eta_L /\beta)\leq \beta\Theta+\beta \pi\big({\eta_L \over\beta}\big).
\ee{todual}
On the other hand, we know (cf. \cite[Lemma 1]{nesterov2009primal}) that $\pi$ is smooth with $\|\nabla \pi\|\leq 1$, and $\nabla \pi$ is Lipschitz-continuous w.r.t. to $\|\cdot\|_*$, i.e.,
\[\|\nabla \pi(z)-\nabla\pi(z')\|\leq \|z-z'\|_*\quad \forall z,z'\in E.
\] As a consequence of Lipschitz continuity of $\pi$, when denoting $\pi_\beta(\eta)=\beta \pi\big({\eta\over\beta}\big)$, we have
\[\pi_\beta(\eta_{j-1}+\xi_j)-\pi_\beta(\eta_{j-1})\leq \|\xi_j\|_*,
\]
so that $\bE\left\{\exp\left([\pi_\beta(\eta_{j})-\pi_\beta(\eta_{j-1})]^2/ s^2\right)\right\}\leq \exp(1)$.
Furthermore,
\[
\pi_\beta(\eta_{j-1}+\xi_j)\leq  \pi_\beta(\eta_{j-1})+\la \nabla \pi_\beta(\eta_{j-1}),\xi_j/\beta\ra +{\|\xi_{j}\|_*^2/\beta},
\]
and, because $\eta_{j-1}$ does not depend on $\xi_{j}$ and $\bE\{\|\xi_j\|_*^2\}\leq s^2$, we get
\[\bE_{j-1}\{\pi_\beta(\eta_{j})-\pi_\beta(\eta_{j-1})\}\leq {s^2/\beta}.
\]
By \cite[Proposition 4.2]{juditsky2008large} we conclude that
random variables $\delta_j=\pi_\beta(\eta_{j})-\pi_\beta(\eta_{j-1})$ satisfy for all $t\geq 0$,
\[
\bE_{j-1}\left\{e^{t\delta_j}\right\}\leq \exp\left(ts^2\beta^{-1}+{\tfrac{3}{4}t^2s^2}\right).
\]
Consequently,
\[
\bE\left\{e^{t\pi_\beta(\eta_L )}\right\}\leq \bE\left\{e^{t\pi_\beta(\eta_{L -1})}\right\}\exp\left({ts^2\beta^{-1}}+{\tfrac{3}{4}t^2s^2}\right)\leq
\exp\left({ts^2L\beta^{-1}}+{\tfrac{3}{4}t^2s^2L}\right).
\]
When substituting the latter bound into \rf{todual}, we obtain for $\beta^2={s^2L /\Theta}$
\be
\bE\left\{e^{t\|\eta_L \|_*}\right\}\leq \exp\left(ts\sqrt{\Theta L }+\tfrac{3}{4}t^2s^2L\right)\quad\forall  t\geq 0.
\ee{almostsg}
To complete the proof of the lemma, it remains to show that  \rf{almostsg} implies \rf{sssg}. This is straightforward. Indeed,  for $\chi\sim\N (0,1),\, \alpha>0$ and  $\zeta=\|\eta_L \|_*$ one has
\begin{align*}
\bE\Big\{e^{\alpha\zeta^2}\Big\}&= \bE\left\{ \bE_{\eta}\left( e^{\sqrt{2\alpha}\zeta \chi}\right)\right\} =
\bE_{\chi}\left\{ \bE\left\{ e^{\sqrt{2\alpha}\zeta \chi}\right\}\right\} \\
    & \le \bE_{\chi}\left\{ \exp\left(\sqrt{2\alpha\Theta L }\,s\chi + \tfrac{3}{2}\alpha Ns^2\chi^2\right)\right\} =
    (1 - 3\alpha Ns^2)^{-1/2}\exp\left\{\frac{\alpha\Theta L  s^2}{1-3\alpha Ns^2}\right\}
\end{align*}
When setting $\alpha=(10 \Theta s^2L )^{-1}$, we conclude that
\[
\bE\Big\{e^{\alpha\zeta^2}\Big\}\leq \exp(1)
\]
due to $\Theta\geq 1/2$.\qed
\subsection{Proof of Theorem \ref{thm:reli}}
 We start with analysing the behaviour of the approximate solution $\wh x^k_{m_0}$  at the stages of the preliminary phase of the procedure.
 \begin{lemma}
 \label{lem:ya1}
Let  $m_0=\lceil 64 \delta^2\rho\nu s(4\Theta + {60}t) \rceil$ ((here $\lceil a\rceil$ stands for the smallest integer greater or equal to $a$), $\gamma = (4\nu)^{-1}$, and let $t$ satisfy $t\geq 4\sqrt{2+\log(m_0)}$.
\par
Suppose that $R\geq 2\delta\sigma_*\sqrt{6\rho s/\nu}$, that initial condition $x_0$ of Algorithms \ref{alg2} and \ref{alg3} satisfies $\|x_0-x_*\|\leq R$, and that at the stage $k$ of the preliminary phase we choose
\be\kappa_k = R_{k-1}\sqrt{\frac{\nu(4\Theta+{60}t)}{\rho s m_0}}
\ee{kappa_k} where  $(R_k)_{k\geq 0}$ is defined recursively:
\[
R_{k+1} = \half {R_k} + \frac{{16}\sigma_*^2\delta^2\rho s}{\nu R_k},\quad R_0=R.
\]
Then the approximate solution $\wh x_{m_0}^k$ at the end of the $k$th stage of the CSMD-SR algorithm satisfies, with probability $\geq 1-4ke^{-t}$
\be
    \|\wh{x}_{m_0}^k-x_*\| &\leq R_k \leq  2^{-k}R + {4}\sigma_*\delta\sqrt{{2}\rho s/\nu}.
\ee{ya2}
In particular, the estimate $\wh{x}^{\ov{K}_1}_{m_0}$ after $\overline{K}_1 = \left\lceil \half \log_2\left(\frac{R^2\nu}{{32}\sigma^2_*\delta^2\rho s}\right) \right\rceil $ stages  satisfies with probaility at least $1-4\ov{K}_1 e^{-t}$
\be\|\overline{x}_{m_0}^{\overline{K}_1}-x_*\|\leq {8}\sigma_*\delta\sqrt{{2}\rho s/\nu}.
\ee{prelim-phase}
\end{lemma}
{\bf Proof of the lemma.}\\
{\bf 1$^o$.} Note that initial point $x_0$ satisfies $x_0\in X_R(x_*)$. Suppose that the initial point $x^k_0=\wh x^{k-1}_{m_0}$ of the $k$th stage of the method satisfy $x_0^k\in X_{R_{k-1}}(x_*)$ with probability $1-4(k-1)e^{-t}$.
In other words, there is a set $\B_{k-1}\subset \Omega$, $\Prob(\B_{k-1})\geq 1-4(k-1)e^{-t}$, such that for all $\ov\omega^{k-1}=[\omega_1;...;\omega_{m_0(k-1)}]\subset \B_{k-1}$ one has $x_0^k\in X_{R_{k-1}}(x_*)$.
Let us show that upon termination of the $k$the stage $\wh x^k_{m_0}$ satisfy $\|x_{m_0}^k-x_*\|\leq R_k$ with probability  $1-4ke^{-t}$.
By Proposition \ref{pr:muprop} (with $h(x)=\kappa_k \|x\|$)
we conclude that for some $\ov\Omega_k\subset \Omega$, $\Prob(\ov\Omega_k)\geq 1-4e^{-t}$, solution $\wh{x}^k_{m_0}$  after $m_0$ iterations of the stage satisfies, for all
for all $\omega^k=[\omega_{(k-1)m_0+1},...,\omega_{km_0}]\in \ov\Omega_k$,
\[
{F}(\wh{x}^k_{m_0})-{F}(x_*) \leq \tfrac{1}{m_0}\left(\nu R^2_{k-1}(4\Theta + {60}t) + {\kappa_k R_{k-1}}\right) + \frac{\sigma_*^2}{\nu}\left({\tfrac{7}{4}}+\tfrac{6t}{m_0}\right).
\]
When using the relationship \rf{eq:lemquad} of Assumption [RSC]  we now get
\be
\|\wh{x}^k_{m_0}-x_*\| \leq \delta\left[\rho s\kappa_k +\frac{R_{k-1}}{m_0} +\frac{\nu R^2_{k-1}}{\kappa_k m_0}\left(4\Theta + {60}t\right)+ \frac{\sigma_*^2}{\nu \kappa_k}\left({\tfrac{7}{4}}+\tfrac{6t}{m_0}\right)\right].
\ee{39}
Note that
 $\kappa_k$ as defined in \rf{kappa_k} satisfies
$\kappa_k\leq R_{k-1}(8\delta \rho s)^{-1},$
while $\kappa_km_0\geq 8\delta(4\Theta+{60}t)R_{k-1} \nu$. Because
$m_0\geq {3840} t$ due to  $\rho\nu\geq 1$ and $\delta\geq 1$, one also has
$\left({\tfrac{7}{4}}+\tfrac{6t}{m_0}\right)\kappa_k^{-1}<{16}\delta\rho s/R_{k-1}$.
When substituting the above bounds into \rf{39} we obtain
\be
\|\wh{x}^k_{m_0}-x_*\| \leq \delta R_{k-1}\left(\tfrac{1}{4\delta} + \tfrac{1}{m_0}\right) +
\frac{{16}\delta^2\rho s\sigma_*^2}{R_{k-1}\nu}\leq \half{R_{k-1}} +\frac{{16}\delta^2\rho s\sigma_*^2}{R_{k-1}\nu}=R_k.
\ee{ya111}
We conclude that $\wh{x}^k_{m_0}\in X_{R_k}(x_*)$ for all $\ov\omega^k\in \B_k=\B_{k-1}\cap \ov\Omega_k$, and \[\Prob(\B_k)\geq \Prob(\B_{k-1})-\Prob(\ov\Omega_k^c)\geq 1-4ke^{-t}.\]
\paragraph{2$^o$.}
Let now $a={16}\delta^2\rho s\sigma_*^2/\nu$, and let us study the behaviour of the sequence
\[
R_{k} = \frac{R_{k-1}}{2}+\frac{a}{R_{k-1}}=:f(R_{k-1}), \quad R_0=R\geq \sqrt{2a}.
\]
Function $f$ admits a fixed point at $R=\sqrt{2a}$ which is also the minimum of $f$, so one has  $R_k\geq \sqrt{2a}$ $\forall k$. Thus,
\[
d_k:=R_{k}-\sqrt{2a}= \frac{R_{k-1}-\sqrt{2a}}{2}+\frac{2a-\sqrt{2a}R_{k-1}}{2R_{k-1}} \leq \half d_{k-1}\leq
2^{-k}d_0\leq 2^{-k}(R-\sqrt{2a}).
\]
We deduce that
$R_k\leq 2^{-k}R_0+\sqrt{2a}
$
which is \rf{ya2}.
Finally,
after running $\overline{K}_1$ stages of the preliminary phase, the estimate $\wh{x}^{\overline{K}_1}_{m_0}$ satisfies
\[
\|\wh{x}_{m_0}^{\ov K_1}-x_*\| \leq {8}\delta\sigma_*\sqrt{{2}\rho s/\nu}.\eqno{\mbox{\qed}}
\]
We turn next to the analysis of the asymptotic phase of Algorithm \ref{alg3}. We assume that the preliminary phase of the algorithm has been completed.
\begin{lemma}
Let $t$ be such that $t\geq 4\sqrt{2+\log(m_1)}$, with $m_1=\lceil 81\delta^2\rho s \nu(4\Theta + {60}t) \rceil$, $\gamma=(4\nu)^{-1}$, and let $\ell_k = \lceil 10\times  4^{k-1}\Theta \rceil$. We set
\[\kappa_k = r_{k-1} \sqrt{\frac{\nu(4\Theta+{60}t)}{\rho s m_1}},\quad r_k=2^{-k}r_0, \quad r_0={8}\delta\sigma_*\sqrt{{2}\rho s/\nu}.
\] Then the approximate solution by Algorithm \ref{alg3} $\wh{x}_{m_1}^k$ at the end of the $k$th stage  of the asymptotic phase
satisfies, with probability $\geq 1-4(\overline{K}_1 +k)e^{-t}$, $\|\wh{x}_{m_1}^k-x_*\|\leq r_k$, implying that
\be \|\wh{x}_{m_1}^k-x_*\|\lesssim \delta^2 \sigma_*\rho s\sqrt{\frac{\Theta \left(\Theta+t\right)}{N_k}}, \ee{asymp-phase-batch}
where $N_k = m_1\sum_{i=1}^k \ell_i$ is the total count of oracle calls for $k$ asymptotic stages.
\end{lemma}
{\bf Proof of the lemma.}
Upon terminating the preliminary phase, the initial condition $x_0=\wh x_{m_0}^{\ov K_1}$ of the asymptotic phase satisfies \rf{prelim-phase}
 with probability greater or equal to $1 - 4 \overline{K}_1 e^{-t}$.
 We are to show that  $\forall k \geq 1$, with probability at least $1 - 4(\ov{K}_1 + k) e^{-t}$,
\[\|\wh{x}^{k}_{m_1} - x_*\| \leq r_k = 2^{-k}r_0,\quad r_0={8}\delta\sigma_*\sqrt{{2}\rho s/\nu}.
\]
The claim is obviously true for $k=0$. Let let us suppose that it holds at  stage $k-1\geq 0$, and let us prove that it also holds at stage $k$.
To this end, we reproduce the argument used in the proof of Lemma \ref{lem:ya1}, while taking into account that now $\ell_{k}$ observations are averaged at each iteration of the CSMD algorithm. Recall (cf. Lemma \ref{lem:sum}) that this amounts to replacing sub-Gaussian parameter $\sigma_*^2$ with $\ov\sigma_*^2=10\Theta\sigma_*^2/\ell_{k}$.
When applying the result of Proposition \ref{pr:muprop} and the bound of \rf{eq:lemquad} we conclude (cf. \rf{39})  that, with probability $1-(\ov K_1+k)e^{-t}$,
\[
\|\wh{x}^k_{m_1}-x_*\| \leq \delta\left[\rho s\kappa_k +\frac{r_{k-1}}{m_1} +\frac{\nu r^2_{k-1}}{\kappa_k m_1}\left(4\Theta + {60}t\right)+ \frac{10\Theta\sigma_*^2}{\nu \kappa_k\ell_k}\left({\tfrac{7}{4}}+\tfrac{6t}{m_1}\right)\right]
\]

By simple algebra, we obtain the following analogue of \rf{ya111}:
\[
\|\wh{x}^k_{m_1}-x_*\| < \delta r_{k-1}\left(\tfrac{2}{9\delta} + \tfrac{1}{{m_1}}\right) +
{10}\frac{4^{-k+1}\delta^2\rho s\sigma_*^2}{r_{k-1}\nu}< \tfrac{r_{k-1}}{4} +\tfrac{r_{k-1}}{4}=r_k.
\]
Observe that upon the end of the $k$th stage we used
$N_k = m_1\sum_{i=1}^{k}\ell_k< 3m_1\Theta \sum_{j=1}^k4^{j-1}\leq  4^k \Theta m_1$ observations of the asymptotic stage. As a consequence,
$4^{-k}<\Theta m_1/N_k$ and
\[r_k= 2^{-k}r_0
\lesssim \delta^2\sigma_*\sqrt{\Theta(\Theta+t) s\nu\rho\over N_k}.\eqno{\mbox{\qed}}
\] 
Assuming that the preliminary phase of Algorithm \ref{alg2} was completed, we now consider the asymptotic phase of the algorithm.
\begin{lemma}\label{ya2222}
Let  $t\geq 4\sqrt{2+\log m_{k}}$, $m_{k}=\left\lceil  4^{k+4}(4\Theta + {60}t)\delta^2\rho s\nu\right\rceil$,
\be
\gamma^k ={r_{k-1}\over {2}\sigma_*}\sqrt{(4\Theta+{60}t)\over {2}m_k}, \quad\kappa_k^2={{5}\sigma_*r_{k-1}\over \rho s}\sqrt{{}\left(4\Theta + {60}t\right)\over m_k}
\ee{gamk2} where
\[r_k:=2^{-k}r_0,\quad r_0={8}\delta\sigma_*\sqrt{{2}\rho s/\nu}.
\]
\par
Then the approximate solution $\wh{x}^k_{m_{k}}$ upon termination of the $k$th asymptotic stage satisfies with probability $\geq 1-4(\overline{K}_1 +k)e^{-t}$
\be \begin{array}{rll}
    \|\wh{x}_{m_{k}}^k-x_*\|\leq 2^{-k}r_0 \lesssim & 2^{-k}\sigma_*\delta\sqrt{\rho s \nu^{-1}}
    \lesssim &\delta^2 \sigma_*\rho s\sqrt{\frac{ \Theta+t}{N_k}}
\end{array} \ee{asymp-phase-nobatch}
where $N_k = \sum_{j=1}^{k}m_{j}$ is the total iteration count of $k$ stages of the asymptotic phase.
\end{lemma}
{\bf Proof of the lemma.}

We are to show that $\forall  k\geq 0$,
$\|\wh{x}^k_{m_k}-x_*\|\leq r_k$ with probability $\geq 1-4(\overline{K}_1+k)e^{-t}$ is true. By Lemma \ref{lem:ya1}, the claim is true for $k=0$ (at the start of the asymptotic phase, the initial condition $x_0=\wh x^{\ov K_1}_{m_0}$ satisfies the bound \rf{prelim-phase}).
We now assume it to hold for $k-1\geq 0$, our objective is to implement the recursive step $k-1\to k$ of the proof.
First, observe that the choice of $\gamma^k$ in \rf{gamk2} satisfies $\gamma^k\leq (4\nu)^{-1}$, $k=1, ...$, so
that Proposition \ref{pr:muprop} can be applied. From the result of the proposition and bound \rf{eq:lemquad} we conclude (cf. \rf{39})  that  it holds, with probability $1-(\ov K_1+k)e^{-t}$,
\[
\|\wh{x}^k_{m_k}-x_*\| \leq \delta\left[\rho s\kappa_k +\frac{r_{k-1}}{m_k} +\frac{r^2_{k-1}\left(4\Theta + {60}t\right)}{\gamma^k\kappa_k {m_k}}
+ {8\frac{\gamma^k\sigma_*^2}{\kappa_k}}\right]
\]
When substituting the value of $\gamma^k$ from \rf{gamk2} we obtain
\[
\|\wh{x}^k_{m_k}-x_*\| \leq \delta\left[\rho s\kappa_k +\frac{r_{k-1}}{m_k} +\frac{{4}
\sigma_*r_{k-1}}{\kappa_k}\sqrt{{2}(4\Theta + {60}t)\over m_k}\right],
\]
which, by the choice of $\kappa_k$ in \rf{gamk2}, results in
 results in\[
\|\wh{x}^k_{m_k}-x_*\|^2 \leq 2\delta^2\left[{10}\rho s \sigma_*r_{k-1}\sqrt{{}4\Theta + {60}t\over m_k}+\frac{r^2_{k-1}}{m_k^2}\right]\leq \frac{r_{k-1}^2}{4}=r_k^2.
\]
It remains to note that the total number $N_k=\sum_{j=1}^k m_j$ of iterations during $k$ stages of the asymptotic phase satisfies $N_k\lesssim 4^k (\Theta + t)\delta^2\rho s\nu$, and $2^{-k}\lesssim \delta\sqrt{(\Theta + t)\rho s\nu
\over N_k}$, which along with definition of $r_0$ implies \rf{asymp-phase-nobatch}. \qed

\paragraph{Proof of Theorem \ref{thm:reli}.} We can now terminate the proof of the theorem.
Let us prove the accuracy bound of the theorem for the minibatch variant of the procedure.

Assume that the ``total observation budget'' $N$ is such that only the preliminary phase of the procedure is implemented.
This is the case when either $m_0\overline{K}_1\geq N$, or $m_0\overline{K}_1< N$ and $m_0\overline{K}_1 + m_1 \ell_1> N$.
The output $\wh{x}_N $ of the algorithm is then the last update of the preliminary phase, and by Lemma \ref{lem:ya1} it satisfies
 $\|\wh{x}_N-x_*\|\leq R2^{-k}$
 where $k$ is the count of completed stages.
In the case of $m_0\overline{K}_1\geq N$ this clearly implies that (recall that $N\geq m_0$) that $k\geq cN/m_0$ and, with probability $\geq 1-4ke^{-t}$
\be
\|\wh{x}_N-x_*\|\lesssim R\exp\left\{-\frac{c' N}{\delta^2\rho s \nu(\Theta +t)}\right\}.
\ee{pbo}
On the other hand, when $m_0\overline{K}_1< N<m_0\overline{K}_1 + m_1 \ell_1$, by definition of $m_1$ and $\ell_1$, one has $N\leq C m_0\overline{K}_1$, so that bound
\rf{pbo} still holds in this case.

Now, consider the case where at least one asymptotic stage has been completed.
When $m_0\overline{K}_1>\frac{N}{2}$ we still have $N\leq C m_0\overline{K}_1$, so that the bound \rf{pbo} holds for the approximate solution $\wh{x}^{(b)}_N$ at the end of the asymptotic stage. Otherwise, the number of oracle calls $N_k$ of asymptotic stages satisfies $N_k\geq N/2$, and by \rf{asymp-phase-batch} this implies that with probability $\geq 1-4(\overline{K}_1+\overline{K}_2)e^{-t}$,
 \[
 \|\wh{x}^{(b)}_N-x_*\|\lesssim \delta^2\sigma_*\rho s \sqrt{\frac{\Theta(\Theta+t)}{N}}.
 \]
To summarize, in both cases, the bound of Theorem \ref{thm:reli} holds with probability at least $1-4(\overline{K}_1+\overline{K}_2)e^{-t}$.

The proof of the accuracy bound for the ``standard'' solution $\wh x_N$ is completely analogous, making use of the bound \rf{asymp-phase-nobatch} of Lemma \ref{ya2222}  instead of \rf{asymp-phase-batch}.
 \qed

\begin{remark}
Theorem \ref{thm:reli} as stated in Section \ref{sec:multi_alg} does not say anything about convergence of $g(\wh x_N)$ to $g(x_*)$. Such information can be easily extracted from the proof of the theorem.
Indeed, observe that at the end of a stage of the method, one has, with probability $1-Cke^{-t}$,
\[
F_{\kappa_k}(\wh x^k)-F_{\kappa_k}\leq \upsilon_k,
\]
or
\[
g(\wh x^k)-g(x_*)\leq \upsilon_k+\kappa_k(\|\wh x^k\|-\|x_*\|)\leq \upsilon_k+\kappa_k\|\wh x^k-x_*\|
\]
where $\wh x^k$ is the approximate solution at the end of the stage $k$. One the other hand, at the end of the $k$th stage of the {\em preliminary} phase one has
$
\|\wh{x}^k-x_*\|\leq R_k\leq 2^{-k}R
$, with
$\kappa_k\lesssim R_k(\delta\rho s)^{-1}\leq 2^{-k}R(\delta\rho s)^{-1}$ and $\upsilon_k\lesssim {4^{-k} R^2\over \delta^2\rho s}$
implying that
\[
g(\wh x^k)-g(x_*)\lesssim \upsilon_k+{R_k^2\over \delta^2 \rho s}
\lesssim (\delta^{-2}+\delta^{-1}){R^2\over \rho s}\exp\left\{-\frac{c}{\delta \rho\nu}{N\over   s(\Theta + t)}\right\}
\]
where $N$ is the current iteration count. Furthermore, at the end of the $k$th {\em asymptotic} stage, one has, with probability $1-(\ov K_1+ k)e^{-t}$,
$
\|\wh{x}^k-x_*\|\leq R_k\lesssim  {\delta^2\sigma_*\rho s}\sqrt{\frac{ \Theta + t}{m_k}},
$
while
$\kappa_k \asymp 2^{-k}\delta\sigma_*(\rho\nu s)^{-1/2}\lesssim \delta\sigma_*\sqrt{\Theta+t\over m_k},
$ and $\upsilon_k\lesssim {\delta^2\sigma_*^2\rho s(\Theta+t)/m_k}$.
As a result, the corresponding $\wh x^k$ satisfies
\[
g(\wh x^k)-g(x_*)\leq \upsilon_k+ \kappa_k\|\wh{x}^k-x_*\|\lesssim(\delta^2+\delta^3)\rho\sigma_*^2s {\Theta+t\over m_k}.
\]
When putting the above bounds together, assuming that at least 1 stage of the algorithm was completed, we arrive at the bound after $N$ steps:
\be
g(\wh x_N)-g(x_*)\lesssim(\delta^{-2}+\delta^{-1}){R^2\over \rho s}\exp\left\{-\frac{c}{\delta^2 \rho\nu}{N\over   s(\Theta + t)}\right\}+(\delta^2+\delta^3)\rho\sigma_*^2s {\Theta+t\over N}
\ee{ghat-g}
with probability $1-(\ov K_1+ \ov K_2)e^{-t}$.
\end{remark}

\subsection{Proof of Proposition \ref{PR:mild}}
{\bf 1$^o$.} Recall that  $\myu$ is $\ol$-Lipschitz continuous, i.e., for  all $t,t'\in \bR^m$
\[
|\myu(t)-\myu(t')|\leq \ol|t-t'|.
\]
As a result, for all $x,x'\in X$,
\begin{align*}
\|\phi[\myu(\phi_{i}^Tx)-\myu(\phi_{i}^Tx')]\|_\infty&\leq \ol \|\phi_{i}\|_\infty |\phi_{i}^T(x-x')|\leq \ol \|\phi_{i}\|^2_{\infty}\|x-x'\|_1\leq \ol\ov\nu^2\|x-x'\|_1,
\end{align*}
so that $\nabla G(x,\omega)=\phi[\myu(\phi^Tx)-\eta]$ is Lipschitz continuous w.r.t. $\ell_1$-norm with Lipschitz constant
$
\L(\omega)\leq \ol \ov\nu^2.$
\paragraph{2$^o$.}
Due to strong monotonicity of $\myu$, 
\bse
g(x)-g(x_*)&=&\int_0^1 \nabla g(x_*+t(x-x_*))^T(x-x_*)dt\nn&=&
\int_0^1 \bE\Big\{\phi [\myu(\phi^T(x_*+t(x-x_*))-\myu(\phi^Tx_*)]\Big\}^T(x-x_*)dt\nn
&\geq &
\int_0^1 \ul \bE\big\{(\phi^T(x-x_*))^2\big\}tdt=
\half \ul \|x-x_*\|^2_\Sigma,
\ese
what is \rf{eq:qmin}.
\paragraph{3$^o$.} The sub-Gaussianity in the ``batchless'' case is readily given by $\nabla G(x_*,\omega_i)=\sigma\phi_i\xi_i$ with
$\|\phi_i\xi_i\|_\infty\leq \|\phi_i\|_{\infty}|\xi_i|\leq \ov\nu\|\xi_i\|_2$ and
\[
\bE\left\{\exp\left({\|\nabla G(x_*,\omega_i)\|_\infty^2\over \sigma^2\ov\nu^2}\right)\right\}\leq e
\]
due to  $\bE\big\{e^{\xi_i^2}\big\}\leq \exp(1)$. Because $\Theta$ variation of the d.-g.f.  $\theta$, as defined in \rf{thel1}, is bounded with $C\ln n$, by Lemma \ref{lem:sum} we conclude that batch observation
\[
H\left(x_*,\omega^{(L)}_i\right)={1\over L}\sum_{\ell=1}^L \nabla G(x_*,\omega^{\ell}_i)={1\over L}\sum_{\ell=1}^L \sigma\phi^\ell_i,\xi^{\ell}_i
\]
is sub-Gaussian with parameter $\lesssim \sigma^2 \ov\nu^2\ln n$.
\paragraph{4$^o$.}
In the situation of Section \ref{sec:rps}, $\Sigma$ is positive definite, $\Sigma\succeq \kappa_\Sigma I$, $\kappa_\Sigma>0$, and condition \ref{z_I}
is satisfied with $\lambda=\kappa_\Sigma$ and $\psi=1$.
Because quadratic minoration condition \rf{qminc} for $g$ is verified with $\mu\geq \ul$ due to \rf{eq:qmin}, when applying the result of Lemma \ref{lem:3.1sparse}, we conclude that Assumption [RSC] holds with { $\delta=1$} and $\rho=(\kappa_\Sigma\ul)^{-1}$.\footnote{We refer to Section \ref{sec:cq} and Lemma \ref{lemm1} for the proof of Lemma \ref{lem:3.1sparse}.}\qed

\section{Properties of sparsity structures}
%
%
\subsection{Sparsity structures}
The scope of results of Section \ref{sec:statement} is much broader than ``vanilla'' sparsity optimization. We discuss here general notion of {\em sparsity structure} which provides a proper application framework for these results.

In what follows we assume to be given a {\em sparsity structure} \cite{juditsky2014unified} on $E$---a family $\cP$ of projector mappings $P=P^2$ on $E$ such that
\begin{itemize}\item[\bf A1.1]
every $P\in \cP$ is assigned a linear map $\ov{P}$ on $E$ such that $P\ov{P}=0$ and a nonnegative weight $\pi(P)$;
\item[\bf A1.2]  whenever $P\in \P$ and $f,g\in E$ such that $\|f\|_*\leq 1$, $\|g\|_*\leq 1$,
\[
\|P^*f+\ov{P}^*g\|_*\leq 1
\]
where for a linear map $Q:\,E\to F$, $Q^*:\,F\to E$ is the conjugate mapping.
\end{itemize}
Following \cite{juditsky2014unified}, we refer to a collection of the just introduced entities and {\em sparsity structure on $E$}.
For a nonnegative real $s$ we set
\[
\cP_s=\{P\in \cP:\pi(P)\leq s\}.
\]
Given $s\geq 0$ we call $x\in E$ {\em $s$-sparse} if there exists $P\in \cP_s$ such that $Px=x$.

Typically, one is  interested in the following ``standard examples'':
\begin{enumerate}
\item ``Vanilla (usual)'' sparsity: in this case $E=\bR^n$ with the standard inner product, $\cP$ is comprised of projectors on all coordinate subspaces of $\bR^n$, $\pi(P)=\rank(P)$, and $\|\cdot\|=\|\cdot\|_1$.
\item Group sparsity: $E=\bR^n$, and we partition the set $\{1,...,n\}$ of indices into $K$ nonoverlapping subsets $I_1,...,I_K$, so that to every $x\in \bR^n$ we associate blocks $x^k$ with corresponding indices in $I_k,\,k=1,...,K$. Now $\cP$ is comprised of projectors $P=P_I$ onto subspaces $E_I=\{[x^1,...,x^K]\in \bR^n:\,x^k=0\, \forall k\notin I\}$ associated with subsets $I$ of the index set $\{1,...,K\}$. We set $\pi(P_I)=\card I$, and define $\|x\|=\sum_{k=1}^K\|x_k\|_2$---{\em block $\ell_1/\ell_2$-norm.}
\item Low rank structure: in this example $E=\bR^{p\times q}$ with, for the sake of definiteness, $p\geq q$, and the Frobenius inner product. Here $\cP$ is the set of mappings $P(x)=P_\ell x P_r$ where  $P_\ell$ and $P_r$ are, respectively, $q\times q$ and $p\times p$ orthoprojectors, $\ov P(x)=(I-P_\ell) x (I-P_r)$, and
 $\|\cdot\|$ is the nuclear norm $\|x\|=\sum_{i=1}^q\sigma_i(x)$ where $\sigma_1(x)\geq \sigma_2(x)\geq ...\geq \sigma_q(x)$ are singular values of $x$, $\|\cdot\|_*$ is the spectral norm, so that $\|x\|_*=\sigma_1(x)$, and  $\pi(P)=\max[\rank(P_\ell), \rank(P_r)]$.
    \par
    In this case, for $\|f\|_*\leq 1$ and $\|g\|_*\leq 1$ one has
    \[
    \|P^*(f)\|_*=\|P_\ell f P_r\|_*\leq 1,\quad \|\ov P^*(g)\|_*=\|(I-P_\ell) g (I-P_r)\|_*\leq 1,
    \]
    and because the images and orthogonal complements to the kernels of $P$ and $\ov P$ are orthogonal to each other, $\|P^*(f)+\ov P^*(g)\|_*\leq 1$.
\end{enumerate}

\subsection{Condition $\bQ(\lambda, \psi)$}\label{sec:cq}
 We say that a positive semidefinite mapping $\Sigma:\,E\to E$ satisfies condition $\bQ(\lambda, \psi)$ for given $s\in \bZ_+$ if for some $\psi,\lambda>0$ and all $P\in \cP_s$ and $z \in E$
\begin{equation}
    \|Pz\|\leq  \sqrt{s/\lambda} \|z\|_{\Sigma}+\|\ov P z\| -\psi \|z\|. \label{condition Qs}
\end{equation}

\begin{lemma}\label{lemm1}
Suppose that $x_*$ is an optimal solution to \rf{eq:pb} such that for some $P\in \cP_s$, $\|(I-P)x_*\|\leq \delta$, and that condition $\bQ(\lambda, \psi)$ is satisfied.
Furthermore, assume that objective $g$ of \rf{eq:pb} satisfies the following minoration condition
\[
g(x)-g(x_*)\geq \mu\big(\|x-x_*\|_{\Sigma}\big)
\]
where $\mu(\cdot)$ is monotone increasing and convex.
Then  a feasible solution $\widehat{x} \in \mathcal{X}$ to 
\eqref{eq:comp} such that
\begin{align*}
    \Prob\left\{F_{\kappa}(\widehat{x}) - F_k(x_*) \leq \upsilon\right\} \geq 1-\epsilon.
\end{align*}
satisfies, with probability at least $1 - \epsilon$,
\be
    \|\widehat{x} - x_*\| \leq {\mu^*\left(\kappa\sqrt{s/\lambda}\right)+\upsilon\over \kappa\psi}+ {2\delta\over \psi}
\ee{lemma1}
where $\mu^*:\,\bR_+\to \bR_+$ is conjugate to $\mu(\cdot)$, $\mu^*(t)=\sup_{u\geq 0}[tu-\mu(u)]$.
\end{lemma}
\paragraph{Proof.}
When setting $z=\wh x-x_*$ one has
\begin{align*}
\wh x&=\|x_*+z\|=\|Px_*+(I-P)x_*+z\|\geq \|Px_*+z\|-\|(I-P)x_*\|\\&\geq
\|Px_*\|+\|\ov Pz\|-\|Pz\|-\delta
\end{align*}
where we used the relation
\[
\|Px_*+z\|\geq \|Px_*\|-\|Pz\|+\|\ov Pz\|
\]
(cf. Lemma 3.1 of \cite{juditsky2014unified} applied to $w=Px_*$). When using condition  $\bQ(\lambda, \psi)$ we obtain
\[
\|\wh x\|\geq \|Px_*\|-\sqrt{s/\lambda} \|z\|_{\Sigma} +\psi \|z\|-\delta,
\]
so that $F_k(\wh x)\leq F_{k}(x_*)+ \upsilon$ implies
\begin{align*}
\kappa\left(\|Px_*\|+\psi\|z\|-\delta\right)&\leq \half[g(x_*)-g(\wh x)]+\kappa \sqrt{s/\lambda}\|z\|_\Sigma+\kappa \|x_*\|+ \upsilon\\
&\leq  -\half \mu(\|z\|_\Sigma)+\kappa \sqrt{s/\lambda}\|z\|_\Sigma+\kappa \|x_*\| + \upsilon\\
&\leq \half \mu^*(2\kappa\sqrt{s/\lambda})+\kappa\|x_*\|+\upsilon,
\end{align*}
and we conclude that
\[
\kappa\psi\|z\|\leq \half \mu^*(2\kappa\sqrt{s/\lambda})+2\kappa\delta+\upsilon
\]
due to $\|x_*\|-\|Px_*\|\leq \|(I-P)x_*\|\leq \delta$. \qed
\par
Note that when $\mu(u)=\tfrac{\mu}{2} u^2$, one has $\mu^*(t)=\tfrac{1}{2\mu}t^2$, and in the case of $\|\cdot\|=\|\cdot\|_1$, with probability $1-\epsilon$,
\[
\|\wh x-x_*\|_1\leq {s\kappa\over \mu\lambda\psi}+ {\upsilon\over \kappa\psi}+{2\delta\over \psi}.
\]
This, in particular, implies bound \rf{quadsp} of Lemma \ref{lem:3.1sparse}.
\begin{remark}
We discuss implications of condition $\bQ(\lambda, \psi)$ and result of Lemma \ref{lemm1} for ``usual'' sparsity in Section \ref{sec:sr} of the paper. Now, let us consider
the case of the low rank sparsity. Let $z\in \bR^{p\times q}$ with $p\geq q$ for the sake of definiteness. In this case, $\|\cdot\|$ is the nuclear norm, and we put
$P(z)=P_\ell z P_r$ where
$P_\ell$ and $P_r$ are orthoprojectors of rank $s\leq q$ such that $\|(I-P)(x)\|=\|x_*-P_\ell x_*P_r\|\leq \delta$.\footnote{E.g., choose $P_\ell$ and $P_r$ as left and right projectors on the space generated by $s$ principal left and right singular vectors of $x_*$, so that
$\|x_*-P_\ell x_* P_r\|=\|(I-P_\ell) x_*(I-P_r)\|=\sum_{i=s+1}^q \sigma_i\leq \delta$.}

Furthermore, for a $p\times q$ matrix $z$ let us put
\[
\sigma^{(k)}(z)=\sum_{i=1}^k\sigma_i(z),\,\,1\leq k\leq q.
\]
With the sparsity parameter $s$ being a nonnegative integer,
\[
\forall (z\in\bR^{p\times q},P\in\cP_s):\quad\|P(z)\|\leq\sigma^{(s)}(z),\,\;\;\|\overline{P}(z)\|\geq\|z\|-\sigma^{(2s)}(z).\footnotemark
\]
\footnotetext{Indeed, let $P\in\cP_s$, so that $\rank(P_\ell)\leq s$ and $\rank(P_r)\leq s$, and $\|P(z)\|=\|P_\ell z P_r\|\leq \sigma^{(s)}(z)$. Since the matrix $\overline{P}(z)$ differs from $z$ by a matrix of rank at most $2s$, by the Singular Value Interlacing theorem we have $\sigma_i(\overline{P}(z))\geq \sigma_{i+2s}(z)$, whence $\|\overline{P}(z)\|\geq \|z\|-\sigma^{(2s)}(z)$.}
and we conclude that
in the present situation condition
\be\sigma^{(s)}(z)+\sigma^{(2s)}(z)\leq \sqrt{s/\lambda}\|z\|_\Sigma+(1-\psi)\|z\|
\ee{elln}
is sufficient for the validity of $\bQ(\lambda,\psi)$. As a result, condition {\rm (\ref{elln})} with $\psi>0$  is sufficient for applicability of the bound of Lemma \ref{lemm1}. It may also be compared to the necessary and sufficient condition of ``$s$-goodness of $\Sigma$'' 
in \cite{recht2011null}:
\[
\exists \psi>0:\; 2\sigma^{(s)}(z)\leq (1-\psi) \|z\|\;\;\forall z\in\mathrm{Ker}(\Sigma).
\]
\end{remark}
\section{Supplementary numerical experiments}\label{sec:suppexp}

\begin{figure}[ht]
    \centering
  \begin{subfigure}[b]{0.5\linewidth}
    \includegraphics[width=1.\linewidth]{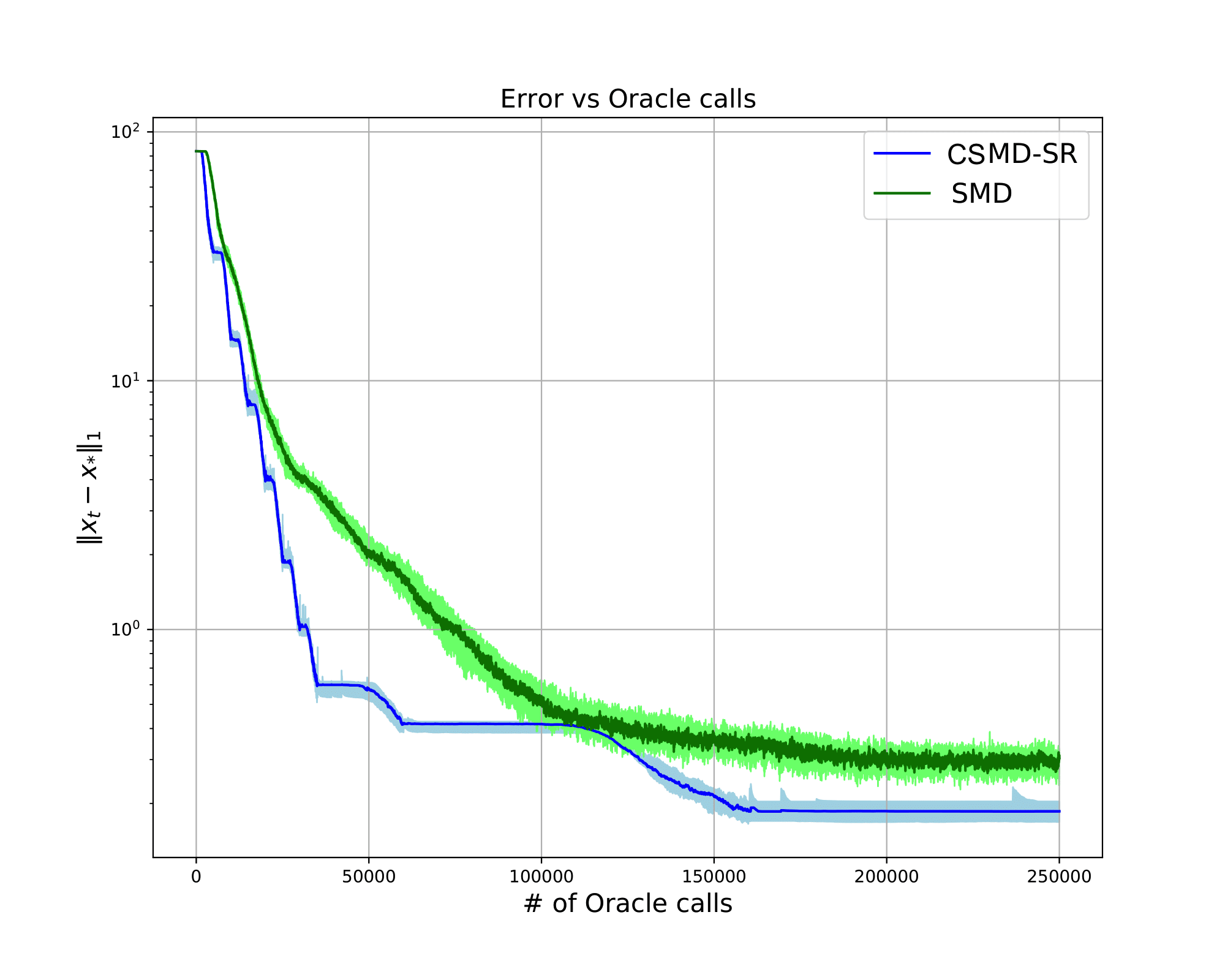}
    \caption{$\kappa_{\Sigma} = 1, \sigma = 0.1, m_0 = 5000$ }
    \label{fig7:a}
    \vspace{3ex}
  \end{subfigure}
  \begin{subfigure}[b]{0.5\linewidth}

    \includegraphics[width=1.\linewidth]{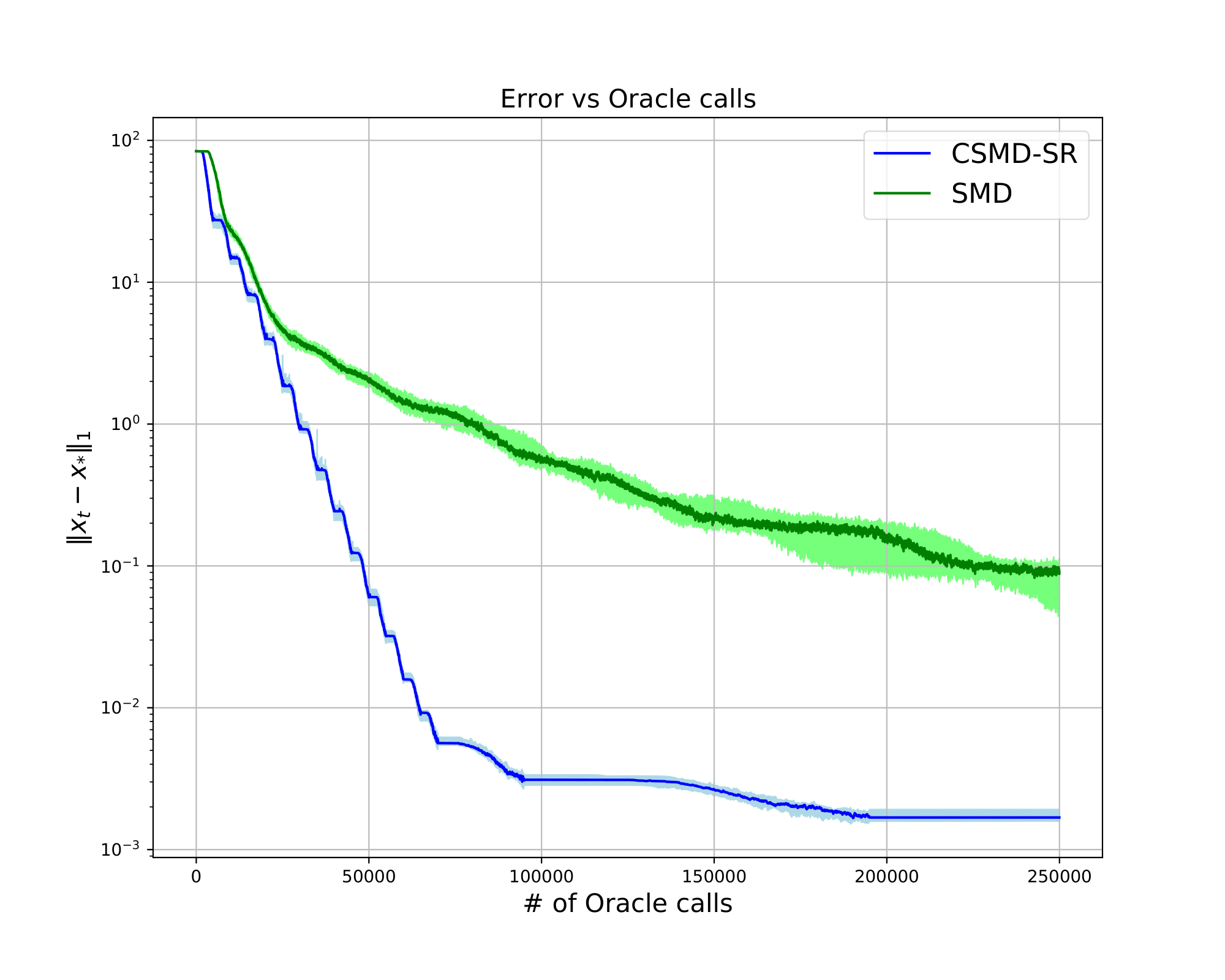}
    \caption{$\kappa_{\Sigma} = 1, \sigma = 0.001, m_0 = 5000$}
    \label{fig7:b}
    \vspace{3ex}
  \end{subfigure}
  \begin{subfigure}[b]{0.5\linewidth}
    \centering
    \includegraphics[width=1.\linewidth]{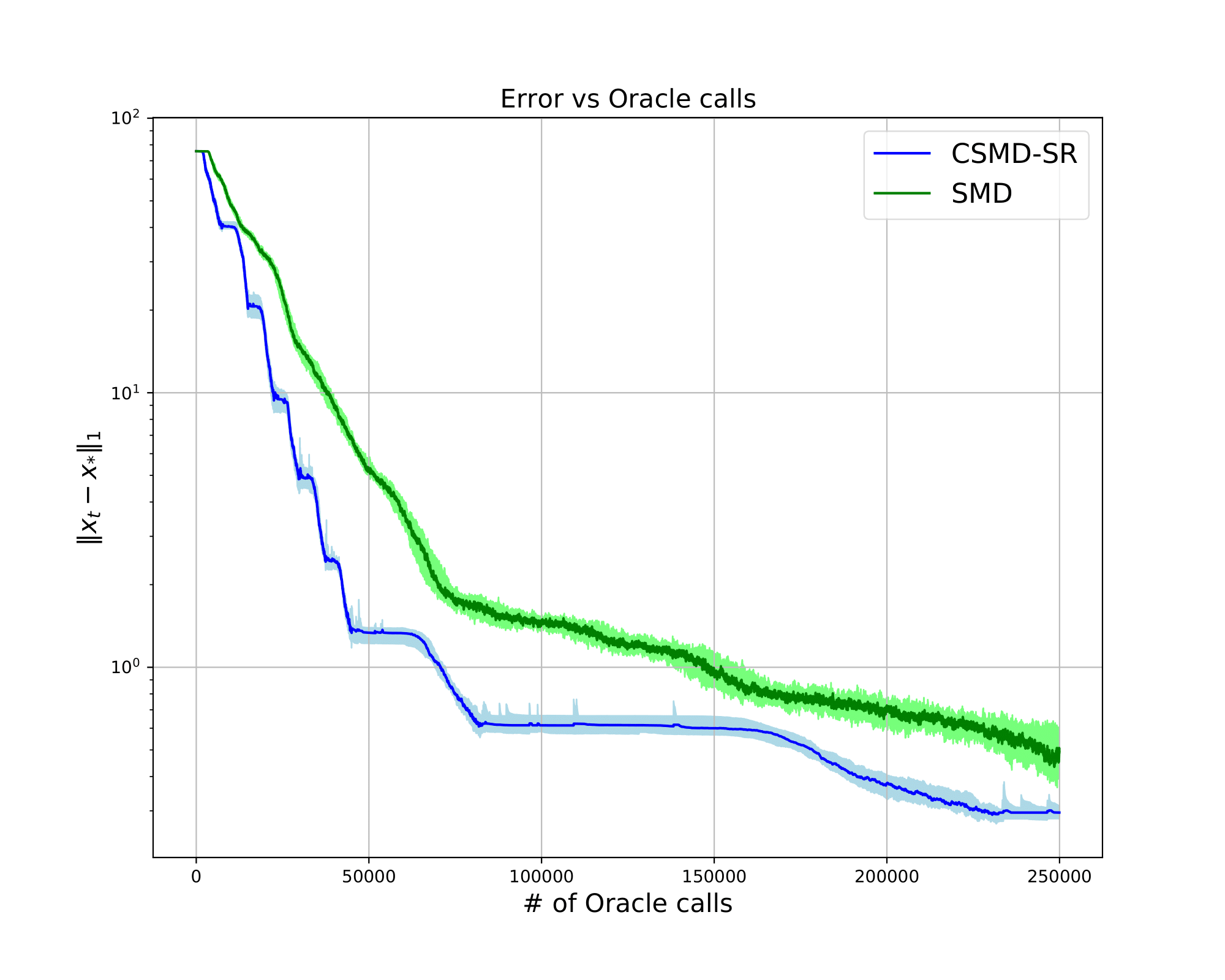}
    \caption{$\kappa_{\Sigma} = 0.1, \sigma = 0.1, m_0 = 7500$}
    \label{fig7:c}
  \end{subfigure}
  \begin{subfigure}[b]{0.5\linewidth}
    \centering
    \includegraphics[width=1.\linewidth]{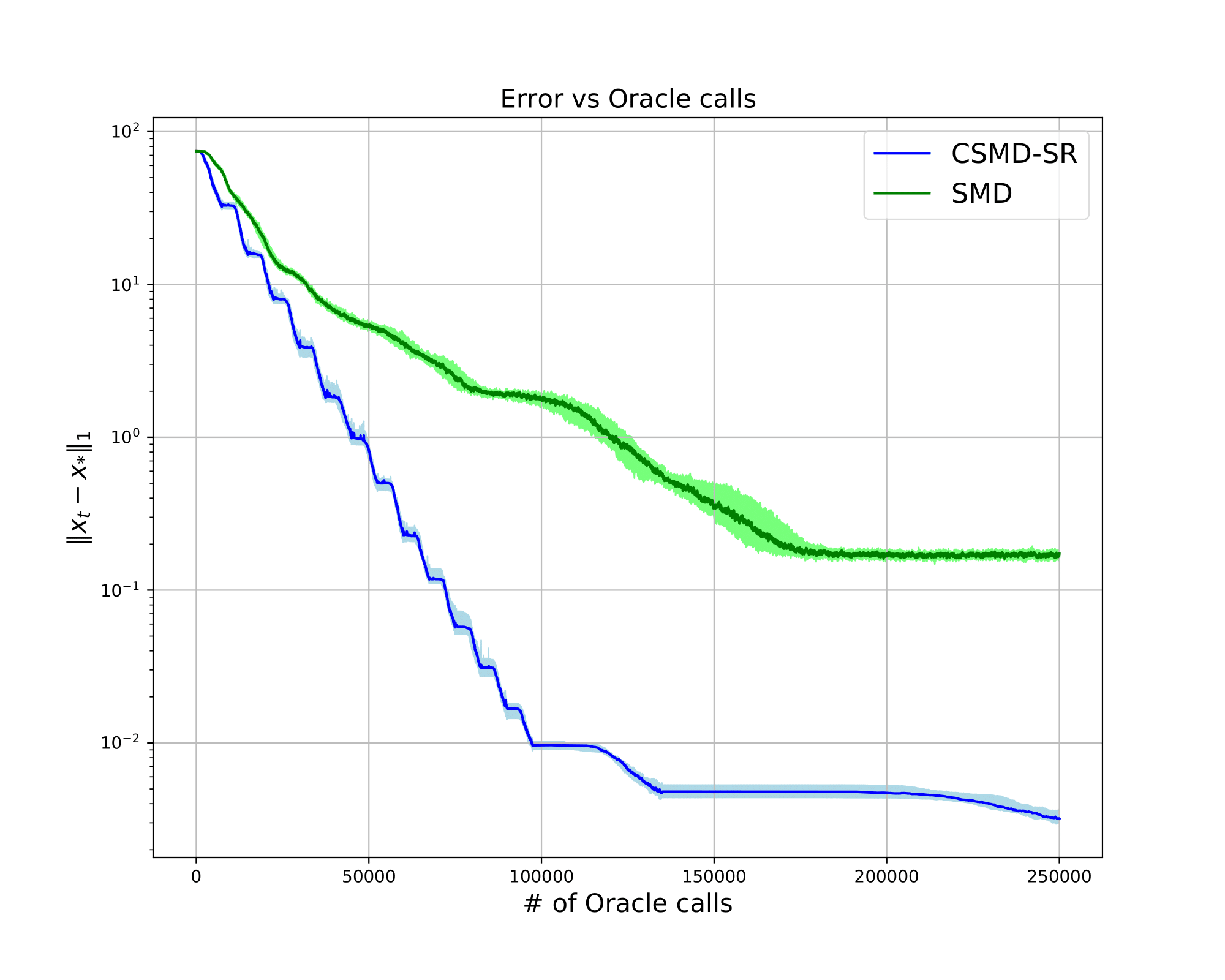}
    \caption{$\kappa_{\Sigma} = 0.1, \sigma = 0.001, m_0 = 7500$}
    \label{fig7:d}
  \end{subfigure}
  \caption{\small CSMD-SR and ``vanilla'' SMD in Linear Regression problem (activation function $\myu(t)=t$); $\ell_1$ error as a function of the number of oracle calls\label{fig3}
}
\end{figure}

\begin{figure}[ht]
    \centering
  \begin{subfigure}[b]{0.5\linewidth}
    \centering
    \includegraphics[width=1.\linewidth]{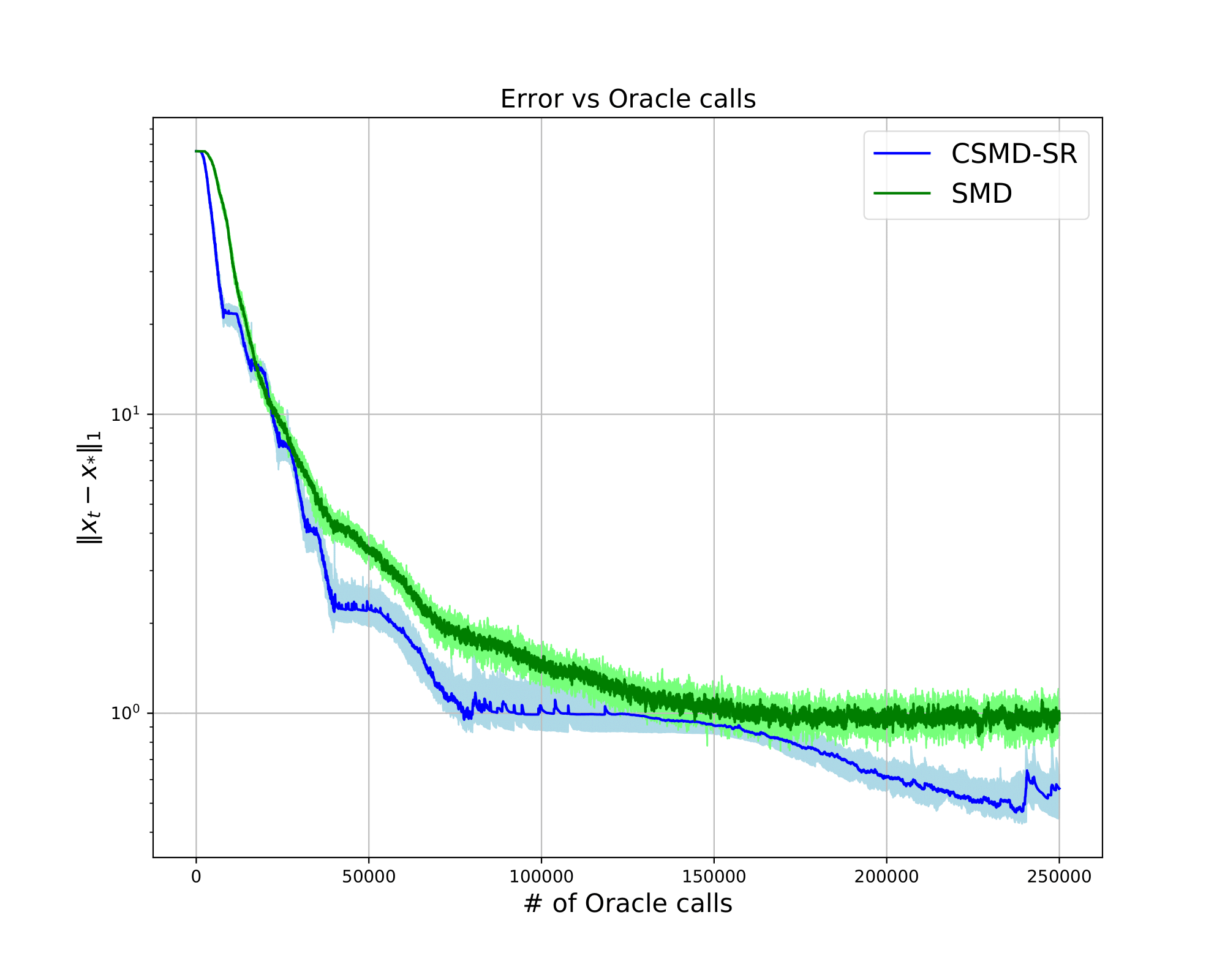}
    \caption{$\kappa_{\Sigma} = 1, \sigma = 0.1, m_0 = 8000$ }
    \label{fig7:a}
    \vspace{3ex}
  \end{subfigure}
  \begin{subfigure}[b]{0.5\linewidth}
    \centering
    \includegraphics[width=1.\linewidth]{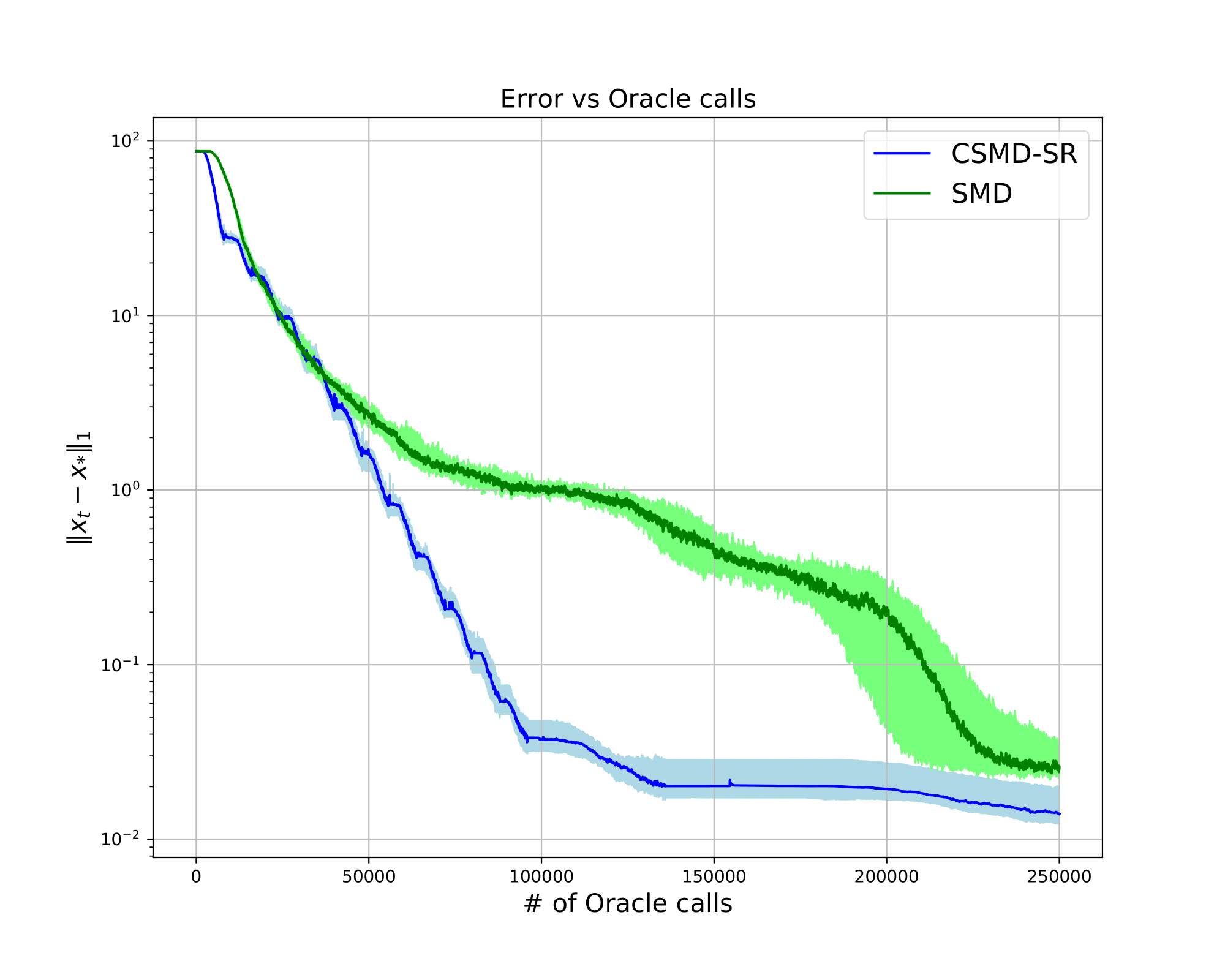}
    \caption{$\kappa_{\Sigma} = 1, \sigma = 0.001, m_0= 8000$}
    \label{fig7:b}
    \vspace{3ex}
  \end{subfigure}
  \begin{subfigure}[b]{0.5\linewidth}
    \centering
    \includegraphics[width=1.\linewidth]{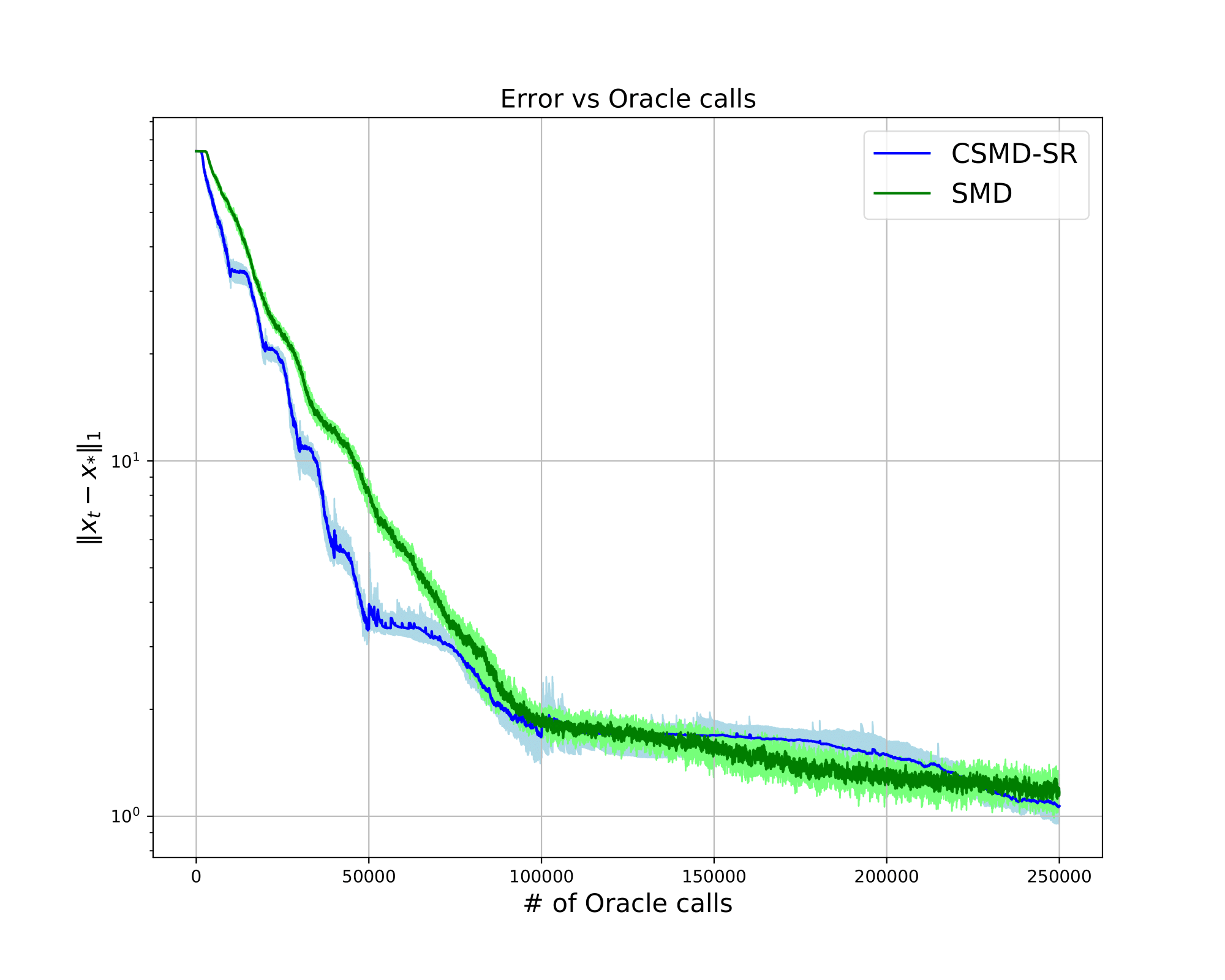}
    \caption{$\kappa_{\Sigma} = 0.1, \sigma = 0.1, m_0=10\,000$}
    \label{fig7:c}
  \end{subfigure}
  \begin{subfigure}[b]{0.5\linewidth}
    \centering
    \includegraphics[width=1.\linewidth]{figs/CSMDvsSMD_kapp_1_1_sig_1_1_m_0_10000_cust.png}
    \caption{$\kappa_{\Sigma} = 0.1, \sigma = 0.001, , m_0=10\,000$}
    \label{fig7:d}
  \end{subfigure}
  \caption{\small CSMD-SR and ``vanilla'' SMD in Generalized Linear Regression problem: activation function $\myu_{1/10}(t)$ ; $\ell_1$ error as a function of the number of oracle calls \label{fig4}}
 \end{figure}

This section complements the numerical results appearing on the main body of the paper. We consider the setting in Section \ref{sec:num} of sparse recovery problem from GLR model observations \eqref{sparselin}.
In the experiments below, we consider the choice  \eqref{ral} of activation function $\myu_{\alpha}(t)$ with values $\alpha = 1$ and $\alpha=1/10$; value $\alpha=1$ corresponds to linear regression with $\myu(t) = t$, whereas when $\alpha=0.1$ activation have a flatter curve with rapidly decreasing with $r$ modulus of strong convexity for $|t|\leq r$. Same as before, in our experiments, the dimension of the parameter space is $n=500\,000$, the sparsity level  of the optimal point $x_*$ is $s = 100$; we use the minibatch Algorithm \ref{alg2-minibatch} with the maximal number of oracle calls is  $ N=250\,000$.
In Figures \ref{fig3} and \ref{fig4} we report results  for $\kappa_{\Sigma}\in\{0.1, 1\}$ and $\sigma\in\{0.001, 0.1\}$; the simulations are repeated 10 times, we trace the median of the estimation error $\|\wh x_i-x_*\|_1$ along with its first and the last deciles against the number of oracle calls.

In our experiments, multistage algorithms exhibit linear convergence on initial iterations. Surprisingly, ``standard'' (non-Euclidean) SMD also converges fast in the ``preliminary'' regime. This may be explained by the fact that iteration $x_i$ of the SMD obtained by the ``usual'' proximal mapping $\Prox(\gamma_{i-1}  \nabla G(x_{i-1},\omega_i),x_{i-1})$ is computed as a solution to the optimization problem with ``penalty'' $\theta(x)=c\|x\|_p^p$, $p=1+1/\ln n$ which results in a  ``natural'' sparsification of $x_i$. As iterations progress, such ``sparsification'' becomes insufficient, and the multistage routine eventually outperforms the SMD.
Implementing the method for ``flatter'' nonlinear activation $\myu(t)$ or increased condition number of the regressor covariance matrix $\Sigma$ requires increasing the length $m_0$ of the stage of the algorithm.

\newpage

\end{document}